\documentclass[11pt]{article}
\usepackage[preprint]{acl}
\usepackage{placeins}
\usepackage{stfloats}
\usepackage{float}
\usepackage{array}
\usepackage{tabularx}
\usepackage{times}
\usepackage{latexsym}
\usepackage[T1]{fontenc}
\usepackage[utf8]{inputenc}
\usepackage{microtype}
\IfFileExists{inconsolata.sty}{\usepackage{inconsolata}}{}
\usepackage{graphicx}
\usepackage{booktabs}
\usepackage{makecell}
\usepackage{adjustbox}
\usepackage{multirow}
\usepackage{amsmath,amssymb}
\usepackage{enumitem}
\usepackage{xspace}
\usepackage{xcolor}
\usepackage{fvextra}
\usepackage{url}
\usepackage{hyperref}
\definecolor{darkblue}{rgb}{0,0,0.5}
\hypersetup{colorlinks=true,citecolor=darkblue,linkcolor=darkblue,urlcolor=darkblue,breaklinks=true}
\setlist{nosep,leftmargin=*}
\newcolumntype{Y}{>{\raggedright\arraybackslash}X}
\newcolumntype{L}[1]{>{\raggedright\arraybackslash}p{#1}}
\newcommand{\vibe}{VIBE\xspace}
\newcommand{\vad}{VAD\xspace}
\newcommand{\affpass}{Affective Passport\xspace}

\newcommand{\score}{\operatorname{Score}}
\newcommand{\generate}{\operatorname{Generate}}

\DeclareMathOperator*{\agg}{agg}
\DeclareMathOperator{\dist}{dist}

\DefineVerbatimEnvironment{vibesnippet}{Verbatim}{fontsize=\scriptsize,breaklines=true,breakanywhere=true}

\title{\vibe: A VAD-Informed Benchmark for Entity-Centered Affective Profiling of Large Language Model Outputs}
\author{
\textbf{Andrei Chetvergov}\thanks{Corresponding author: \texttt{chetvergov-as@ranepa.ru}} \quad
\textbf{Alexander Evseev} \quad
\textbf{Timofei Sivoraksha} \quad
\textbf{Stepan Ukolov} \\
\textbf{Mikhail Solovev} \quad
\textbf{Danil Sazanakov} \quad
\textbf{Sergey Bolovtsov}
}

\begin{document}
\raggedbottom
\maketitle

\begin{abstract}
Large language models routinely describe socially salient targets---political figures, countries, religions, organizations, historical events, and social groups---encoding affective framing alongside factual content: a target may appear favorable or threatening, calm or conflictual, powerful or vulnerable. Existing work captures parts of this space through sentiment, favorability, and emotion benchmarks, but none combines target-directed VAD attribution, an explicit scorer contract, and a passport reporting format. We introduce \vibe, a benchmark for entity-centered affective profiling of LLM outputs in Valence\allowbreak-Arousal\allowbreak-Dominance (\vad) space. Its core contribution is a measurement contract: \vibe separates generation from external scoring, distinguishes scalar favorability, response-level \vad, and target-directed \vad, and reports profiles through an \affpass. Three empirical layers support the contract. H1 shows scalar favorability does not subsume arousal and dominance: valence findings are cross-validated ($r_V{=}0.944$ judge--human, $r_V{=}0.954$ inter-scorer); arousal and dominance are single-scorer directional estimates---not point-precise---consistent with known inter-annotator difficulty on these axes ($r_A^{hh}{=}0.495$, $r_D^{hh}{=}0.702$ among human annotators). H2 shows whole-response and target-directed \vad are different contracts: the same text can carry one affective tone overall while representing the named target differently. H3 is a protocol-drift diagnostic: elicitation conditions shift profiles, motivating context metadata in every affective report. These results motivate entity-centered affective profiling as a documented practice: profiles should be released with scorer identity, coverage, protocol, and interpretation limits.

\end{abstract}

\begin{figure}[t]
\centering
\includegraphics[width=\linewidth]{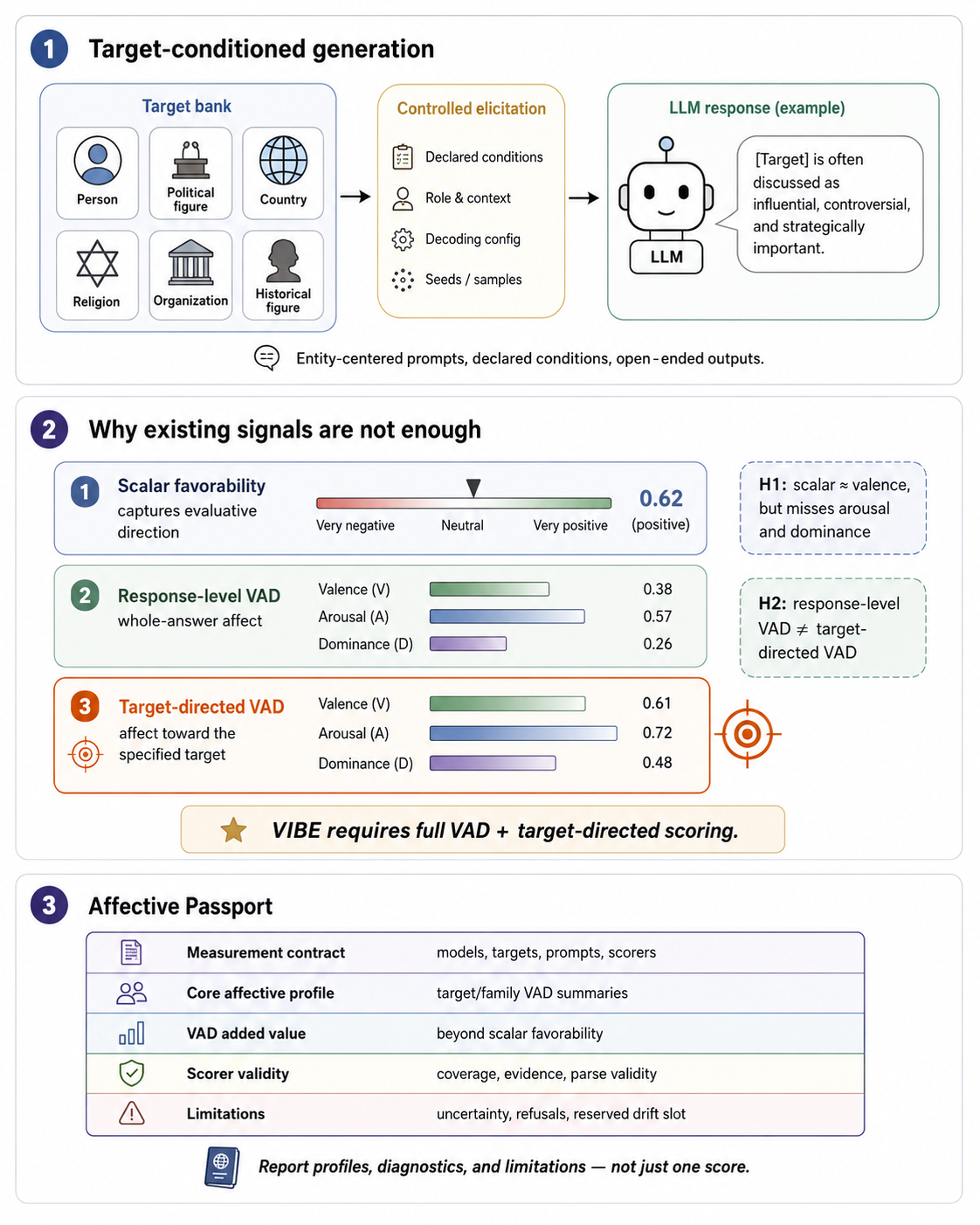}
\vspace{-1.4ex}
\caption{\vibe research framework. A target bank of socially salient entities feeds a controlled generation layer; external scorers apply three complementary scoring contracts (scalar favorability, response-level \vad, and target-directed \vad); evidence layers support H1--H3 (irreducibility, target-directed scoring, protocol drift); the output is an \affpass reporting target-directed affective profiles, scorer-validity diagnostics, coverage, and interpretation limits.}
\label{fig:vibe-contract}
\vspace{-1.2ex}
\end{figure}

\section{Introduction}
Large language models (LLMs) interpret the social world through affective framing—presenting targets as safe or dangerous, powerful, or vulnerable—rather than merely stating facts. While prior work shows that target-conditioned behavior is measurable \citep{buyl2026,bang2024,plaza2024divine,hamborg2021,dufraisse2023} and highly sensitive to evaluation protocols \citep{roettger2024,zheng2023,zeng2024}, standard scalar or categorical metrics miss crucial nuances. A scalar monitor, for instance, passes both Rohingya (favorability$=0.93$) and Philipp Lenard ($0.50$), entirely missing the extreme low-dominance suffering framing of the former and the high-arousal hidden-negative framing of the latter.

We introduce \vibe, a \vad-informed benchmark for entity-centered affective profiling of LLM outputs. Its core contribution is a \emph{measurement contract} that jointly fixes: (i) the measured object (affect toward a named target), (ii) the scorer interface (continuous \vad with explicit scorer identity), (iii) the elicitation protocol, and (iv) the reporting boundary. Rather than claiming to measure internal model emotions, \vibe quantifies affective properties in generated text, reporting them via an \emph{\affpass}.

\paragraph{Contributions.}
\vibe delivers:
\begin{itemize}
    \item \textbf{A measurement contract} separating generation from scoring, with explicit scorer identity and interpretation limits.
    \item \textbf{Quantified substitution costs} across 2{,}613 targets and six models: scalar favorability ($r{=}0.875$ with valence) misses arousal and dominance, and response-level \vad diverges from target-directed \vad (mean L2 $0.240$). 
    \item \textbf{Protocol drift quantification} showing that situation framing shifts profiles far more than model identity ($\eta^2_{\text{family}}{=}0.141$ vs $\eta^2_{\text{model}}{}0.010$), proving that cross-model comparisons require protocol metadata.
    \item \textbf{The \affpass}, a condition-explicit artifact tying every profile to its scorer, protocol, coverage, and limits.
\end{itemize}
\section{Related Work}
\paragraph{Entity-centered sentiment, favorability, and bias.}
The closest prior line evaluates model responses toward specific targets. \citet{buyl2026} score open-ended descriptions of political persons using scalar favorability applied to the whole response; \citet{bang2024} derive entity stance from document-level framing; target-sentiment datasets such as NewsMTSC and MAD-TSC score over a sentence rather than conditioning on the named target \citep{hamborg2021,dufraisse2023}. The shared pattern is that scalar favorability or whole-response labels proxy entity-level affect---this is not a constructed straw-man but the documented practice in each of these lines---and \vibe's H1 and H2 quantify the substitution cost empirically for the first time. \citet{plaza2024divine} show religious targets require attention to stereotypes and emotion representation. Bias benchmarks broaden target coverage \citep{smith2022,parrish2022,nagireddy2024,huangxiong2024,jin2025socialbias} but measure downstream harms and QA behavior rather than continuous target-directed affective profiles.

\paragraph{Dimensional affect and emotion benchmarks.}
\vad provides a compact continuous affect space: valence captures evaluative direction, arousal captures activation, tension, urgency, or emotional intensity, and dominance captures control, agency, power, vulnerability, or dependence. This representation is grounded in affect theory and resources such as the circumplex model, Warriner norms, NRC \vad, and EmoBank \citep{russell1980,warriner2013,mohammad2018,buechel2017}. Emotion NLP datasets and LLM emotion benchmarks evaluate categorical emotions, intensity, dialogue emotion, emotional intelligence, empathy, and role fidelity \citep{demszky2020,mohammad2017wassa,poria2019,sabour2024,huang2024emotionbench,feng2025}. \vibe differs by making the target representation itself the measurement object: the question is how generated text affectively represents a target, rather than whether a model can perform emotional reasoning.

\paragraph{Robustness, persona drift, and evaluator dependence.}
Value and opinion evaluations of LLMs are sensitive to prompt wording, response format, language, role, and persona \citep{roettger2024,moore2024,faulborn2025,liu2024persona,hu2024persona}. Prompt-robustness frameworks and behavioral testing motivate multi-condition evaluation rather than single-template claims \citep{kim2024multiprompt,zhao2024posix,ribeiro2020}. LLM-as-judge work shows that evaluators can be useful but biased, brittle, and in need of calibration \citep{zheng2023,zeng2024,lin2025calibration,chen2024humans,shi2025judging,semeval2025task11}. These findings motivate a central \vibe design choice: scorer configuration and protocol drift must be reported, not hidden.

\paragraph{Measurement and documentation.}
\vibe is framed as a measurement contribution. Construct validity, proxy choice, and interpretation boundaries are central to fairness measurement \citep{jacobs2021,selbst2019}. Data statements, model cards, and datasheets motivate reporting artifacts with provenance, intended use, limitations, and evaluation conditions \citep{bender2018,mitchell2019,gebru2021}. \vibe adapts this logic to target-directed affective profiling through the \affpass.

\section{Task Formulation}
Let $M$ be a set of target models, $T$ a target bank, $P$ prompt templates, $L$ languages, $R$ role or persona settings, $C$ contexts, $K$ generation samples or seeds, $S$ scorers, and $A=\{v,a,d\}$ the \vad axes. A target model generates
\begin{equation}
 y_{m,t,p,l,r,c,k}=\generate(m,t,p,l,r,c,k),
\end{equation}
where $t\in T$ is a named or entity-like socially salient target. A scorer $s$ then returns a target-conditioned score record
\begin{equation}
 x_{m,t,p,l,r,c,k,s}=\score_s(t,y_{m,t,p,l,r,c,k}),
\end{equation}
which includes \vad values and metadata such as confidence, evidence spans, target coverage, refusal/abstention flags, and parse validity. The core profile tensor is
\begin{equation}
 \mathbf{X}[m,t,p,l,r,c,k,s,a],\quad a\in A.
\end{equation}
For a core regime $\mathcal{R}_0$ (e.g., neutral prompt, default role, fixed language, fixed decoding, declared scorer), the target-level profile is
\begin{equation}
 \mu_m(t)=\agg_{(p,l,r,c,k,s)\in\mathcal{R}_0}\mathbf{X}[m,t,p,l,r,c,k,s,:].
\end{equation}
A protocol-drift quantity compares a profile under condition $q$ with the core profile:
\begin{equation}
 \Delta_m(t,q)=\dist\bigl(\mu_m(t\mid q),\,\mu_m(t\mid \mathcal{R}_0)\bigr).
\end{equation}
These records define reported measurements of LLM-generated text---not internal model emotions, beliefs, or general text corpora. Downstream analyses of bias, stigma, or geopolitical framing require comparisons over targets, target families, languages, models, roles, or scorers.

\section{VIBE Framework}
\paragraph{Target ontology and bank construction.}
The benchmark targets socially salient entities across 13 families (person, political\_person, historical\_figure, country, organization, religion, ideology, social\_group, geopolitical\_event, historical\_event, technology, cultural\_symbol, abstract\_phenomenon). The entity bank (v2) was built via SPARQL queries to Wikidata, using \texttt{instance\_of} (P31) constraints per family, a sitelinks threshold for encyclopedic coverage, and Wikipedia pageview salience (12-month window; sitelinks fallback). After deduplication by Wikidata QID, the final bank contains 2{,}613 unique targets (raw: 2{,}846; 233 removed). Each record stores the Wikidata QID, multilingual labels, sensitivity label, and provenance fields. Of the 2{,}613 targets, 40\% carry the \texttt{standard} sensitivity label and 60\% are flagged as politically, geopolitically, socially, or historically sensitive---deliberately skewed toward contested targets where affective profiling matters most. The bank does not overrepresent negative polarity: political persons, ideologies, and geopolitical events ($39\%$ of targets) span valence from near-zero (Khmer Rouge) to near-one (Mahatma Gandhi). Family and sensitivity distributions are in Appendix~\ref{app:model-ids}. The reported experiments use these 2{,}613 targets across H1 and H2, plus external open-description rows for comparability.

\paragraph{Elicitation conditions.}
The full design supports multiple target-conditioned prompt templates and role settings. The empirical sections use a focused subset: brief evaluative prompting as the primary contract, open-description controls for comparability, and selected protocol factors for drift diagnostics. The core profile fixes these conditions; drift analyses intentionally vary them.

\paragraph{Scoring modes.}
\vibe uses three scoring contracts. \textit{Scalar target favorability} scores the target on a single positive--negative dimension. \textit{Response-level \vad} scores the whole response without target visibility. \textit{Target-directed \vad} scores the affect expressed specifically toward the target and is the core \vibe mode. The distinction is load-bearing: a response about a war, religion, or political figure may contain sadness about victims, caution about controversy, or moral condemnation of actions, and response-level scoring can conflate these signals with affect toward the target itself. H2 (Section~\ref{sec:h2}) quantifies this difference.

\paragraph{Scale conventions.}
The repository scorer contracts use a raw $1$--$9$ scale with midpoint $5$ and a normalized $[-1,1]$ convention. The H1 and H2 reports in this paper use a $[0,1]$ scale with neutral point $0.5$; conversion tables to the other two scales are listed in the appendix. Future \vibe releases should record \texttt{scale\_version} alongside every score to prevent silent re-interpretation.

\section{Data and Implementation}
\label{sec:data}
\paragraph{Datasets.}
Primary H1/H2 results come from 2{,}613 socially salient targets (13 families) crossed with six instruction-tuned generators under brief evaluative prompting, yielding 15{,}678 scored generation rows. The target list is released as a paper artifact. H2 starts from the same generation-level response/target score pairs and reports 15{,}626 paper-grade pairs after eligibility filtering. The 52 excluded pairs are documented by low target coverage, low target-prompt agreement or human-review flag, refusal or abstention, or missing evidence. The open-descriptive control reuses the same target universe with 15{,}671 rows. Buyl replication supplies 129{,}181 external rows (Section~\ref{sec:h1}, Step~3). H3 uses a compact score file with 342{,}779 rows across H3.1--H3.5, including an H3.2 instruction-language slice under the same evaluative-stance style. H3 protocol items are \vibe-owned controlled extensions inspired by emotion and dialogue benchmarks; they are not direct replications of external benchmark rows.

\paragraph{Target models and generation.}
The target-model manifest contains six instruction-tuned generators from Russian, European, Google-family, IBM-family, and Qwen-family model lines; Appendix~\ref{app:model-ids} lists the display names and run identifiers used for reproducibility. Default generation settings are temperature $0.0$, top-$p$ $0.95$, seed $13$, max tokens $512$, and parallelism $8$ via an OpenAI-compatible API layer. External scorer configs use OpenAI-compatible judge adapters with temperature $0.0$ and JSON-object output.

\paragraph{Scorer assignment.}
Primary H1/H2 target-directed \vad and scalar favorability use a Qwen3.6 LLM judge for budget and throughput on five of six generators. To limit same-family self-judgment, outputs from the Qwen generator (\texttt{qwen3\_6\_35b\_a3b}) are scored with a Gemma-4 judge instead, and Gemma-generator outputs are scored with Qwen (symmetric swap). The headline H1/H2 tables therefore mix two primary judges by design; Section~\ref{sec:h1} reports a GPT-4o-mini cross-check separately for the five-model Qwen-primary slice and for the Qwen-generator slice scored by Gemma. We do not compare scalar favorability across judges. Every scored row stores \texttt{scorer\_model} so downstream artifacts can filter by judge identity.

\paragraph{Why LLM judges instead of lexicon or \vad regression?}
We tested cheaper scorers and rejected them as \emph{primary} instruments. \textbf{NRC-VAD} lexical means \citep{mohammad2018} are not target-directed; on our 15{,}678 brief evaluative generations they track the LLM judge weakly (NRC-VAD valence vs.\ Qwen judge: $r\approx 0.60$; cross-scorer Qwen--GPT agreement: $r=0.95$; mean \vad{} distance $\approx 0.72$). \textbf{Pretrained \vad regressors} \citep{vibe008} lack target-visible attribution and are trained on word- and sentence-level supervision that mismatches our multi-sentence evaluative paragraphs. \vibe uses contract-aware LLM judges for H1/H2/H3; lexicon and regression tools serve as offline sanity checks in the artifact bundle.

LLM judges introduce their own family-specific biases \citep{chen2024humans,shi2025judging,zheng2023,zeng2024,lin2025calibration}; the scorer-swap policy and GPT-4o-mini cross-check described above are the primary mitigations.

\section{Experiments and Hypotheses}
\label{sec:experiments}
The current empirical scope of \vibe is organized around three hypotheses, each grounded in a distinct measurement question. A compact status table is provided in the appendix.

\paragraph{H1: scalar favorability vs.\ \vad decomposition.}
H1 asks whether scalar favorability is sufficient for target-directed affective representation. H1.1 predicts alignment between scalar favorability and target-directed valence. H1.2--H1.5 test irreducibility: arousal and dominance should vary within scalar bins, retain absolute residual signal from neutrality, and reveal hidden affect even when valence is near-neutral. H1.6 tests whether these patterns vary across target families. Results are reported in Section~\ref{sec:h1} on three elicitation settings---brief evaluative prompting (primary \vibe contract), open descriptive prompting (control), and Buyl replication---using six models on the \vibe-owned tracks.

\paragraph{H2: response-level vs.\ target-directed scoring.}
H2 asks whether whole-response \vad tone differs from target-directed \vad on the same text. Paired records (response-level and target-directed, same scorer) are compared by Euclidean distance and per-axis deltas on 15{,}626 paper-grade generations (Section~\ref{sec:h2}).

\paragraph{H3: protocol-drift diagnostic.}
H3 asks whether target-directed profiles should travel with their elicitation conditions---a proof-of-need for passport metadata, not a full robustness benchmark. Five \vibe-owned protocol families are tested (H3.1 situation/factor; H3.2 instruction language; H3.3 task regime; H3.4 role-play depth; H3.5 dialogue-topic shell), with situation framing predicted to dominate. H3.2 varies the instruction language across seven languages (en, ru, fr, es, zh, ar, ja); models respond in the instruction language, producing multilingual outputs for the same targets. Results in Section~\ref{sec:h3}.

\section{Results: H1 \vad Irreducibility}
\label{sec:h1}
The H1 layer asks whether scalar favorability is sufficient to recover a target-directed \vad profile. We test this on a three-step elicitation ladder (Table~\ref{tab:h1-elicitation-comparison}): (i)~\textbf{brief evaluative prompting} on the \vibe entity bank---models are asked for a short evaluative stance toward the target rather than an encyclopedic description; (ii)~\textbf{open descriptive prompting} on the same bank---a neutral ``tell me about the target'' template in the style of open-description benchmarks; and (iii)~\textbf{Buyl replication} at scale for external comparability. All tracks pair scalar favorability with the same target-directed \vad scorer on a $[0,1]$ scale (neutral point $0.5$); metric definitions are in Appendix~\ref{app:h1-metrics}. Six instruction-tuned models are scored on the \vibe-owned tracks.

\paragraph{Reading these results.}
Valence findings are \emph{cross-validated}: judge--human $r_V{=}0.944$ and inter-scorer $r_V{=}0.954$ confirm direction and ordering. Arousal and dominance are \emph{single-scorer directional estimates} bounded by construct difficulty---human annotators agree less on these axes ($r_A^{hh}{=}0.495$, $r_D^{hh}{=}0.702$, $r_V^{hh}{=}0.798$; 23 distinct mapping points from 28 emotion labels impose a granularity ceiling on human-side precision). Report A/D as ordinal tendencies, not point estimates.

\paragraph{Step 1: brief evaluative prompting (primary \vibe contract).}
All 15{,}678 target-conditioned generations have paired scalar favorability and target-directed \vad scores ($100\%$ join coverage). Scalar--valence alignment tightens ($r=0.9491$), yet H1.2--H1.5 remain supported: within-bin arousal/dominance variation ($0.1806$), dominance residual ($0.2786$), and arousal/dominance signal in near-neutral-valence rows ($41.19\%$ under primary scorer; Table~\ref{tab:h1-evaluative-stance}; single-scorer estimate---see Appendix~\ref{app:hidden-affect-quantization}). Cross-model disagreement is substantial (Appendix Figures~\ref{fig:h1-disagreement-map},~\ref{fig:h1-radar-cases}); score quantization is expected from ordinal-granularity LLM output and does not affect the co-occurrence analysis (Appendix Figure~\ref{fig:h1-vad-distributions}).

\paragraph{Step 2: open descriptive prompting (control on the \vibe bank).}
Here the generation prompt follows the open-description family used in Buyl-style benchmarks: neutral encyclopedic coverage of the target. On 15{,}671 rows (six models), irreducibility persists (Pearson $r=0.8749$; hidden-affect share $26.68\%$; Table~\ref{tab:h1-vibe-own} in the appendix). This step separates ``Buyl subsample only'' from ``open-description elicitation in general'' and shows that irreducibility is not specific to the brief evaluative wording.

\paragraph{Step 3: Buyl replication (external scale).}
The Buyl-overlap track provides 129{,}181 usable scored rows (99.9930\% coverage) under the same open-description prompt family on their political-person subsample. Irreducibility is already visible before any \vibe-specific prompt design: scalar favorability aligns with valence ($r=0.7586$) but does not subsume arousal and dominance (within-bin std.\ $0.1870$; arousal/dominance signal in neutral-valence rows $42.56\%$ under primary scorer). Appendix~\ref{app:h1-diagnostics} provides the full visual diagnostics.

\begin{figure*}[t]
\centering
\includegraphics[width=\linewidth]{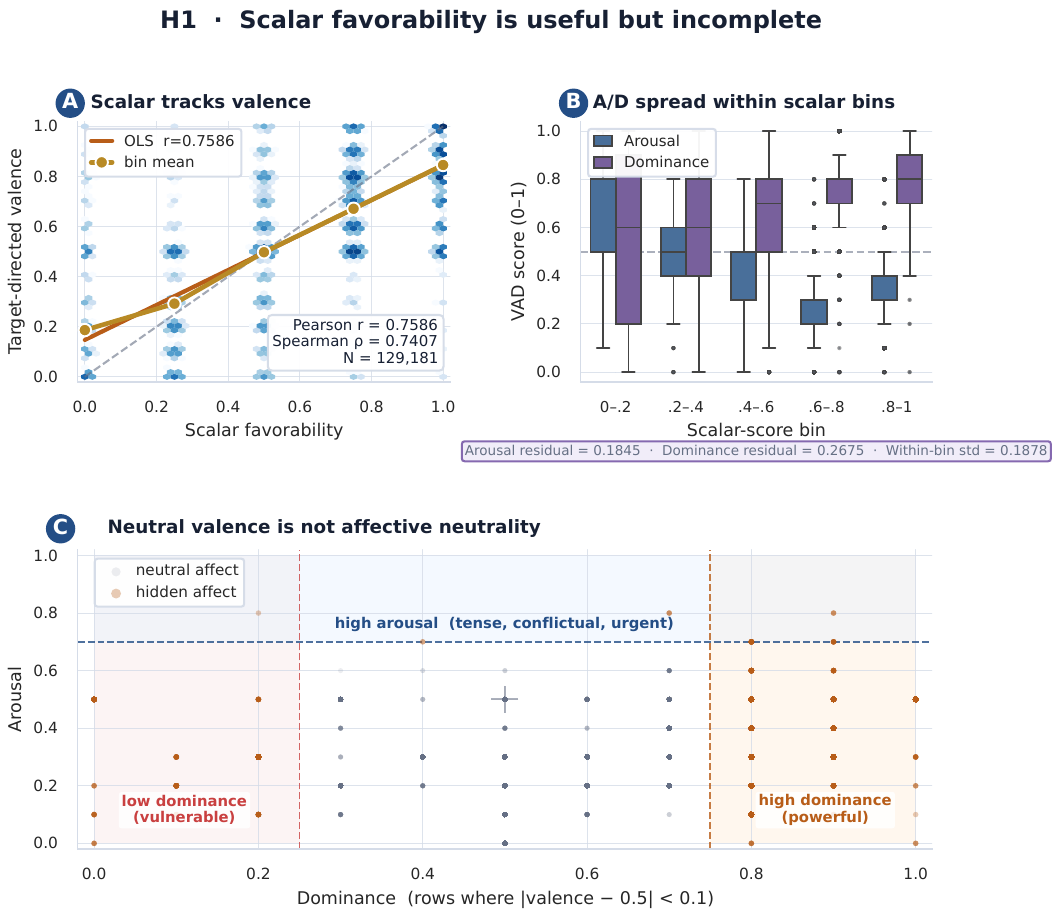}
\vspace{-1.8ex}
\caption{H1 visual story (Buyl replication track, $n=129{,}181$ rows). Panel~A: scalar favorability vs.\ target-directed valence (Pearson $r=0.7586$); the OLS line and bin means confirm strong but imperfect alignment. Panel~B: arousal and dominance distributions per scalar-score bin; similar favorability ranges contain substantial \vad variation. Panel~C: near-neutral-valence rows in arousal--dominance space; 42.56\% of rows with valence $\in[0.45,0.55]$ still exhibit non-neutral arousal or dominance under the primary scorer (orange points), illustrating residual affect signal that a scalar score cannot recover.}
\label{fig:h1-story}
\vspace{-1.0ex}
\end{figure*}

\paragraph{Validation checks.}
Headline H1 metrics follow the mixed judge policy (Section~\ref{sec:data}). Qwen and the swap scorer agree on valence ($r{=}0.954$) but less strongly on arousal ($r{=}0.418$) and dominance ($r{=}0.573$) across $13{,}065$ paired rows (Appendix Table~\ref{tab:h1-scorer-agreement}); the headline pattern holds on the five non-Qwen generators. Full inter-scorer agreement is in Appendix~\ref{app:inter-scorer}.

\paragraph{Human calibration: directional structure, not point-level agreement.}
On 325 targets (four annotations each), judge--human agreement at target level is $r_V{=}0.944$, $r_A{=}0.773$, $r_D{=}0.863$ (Qwen3.6; mean L2 $0.195$). Human inter-annotator agreement is $r_V^{hh}{=}0.798$, $r_A^{hh}{=}0.495$, $r_D^{hh}{=}0.702$ (pairwise, four annotators)---the emotion-label-to-\vad mapping (28 labels $\rightarrow$ 23 distinct \vad points) imposes a granularity ceiling on human precision \citep{demszky2020,warriner2013}. Judge $r_A{=}0.773$ exceeds the human-human floor ($r_A^{hh}{=}0.495$), confirming A/D uncertainty reflects construct difficulty \citep{buechel2017,lin2025calibration}, not a model failure; a perturbation simulation confirms directional agreement holds in $100\%$ of draws (Appendix~\ref{app:proxy-sensitivity}). Full results: Appendix Table~\ref{tab:h1-human-judge-validation}.

\paragraph{Hidden affect: operational definition and example.}
H1.5 operationalises \emph{residual affect}: near-neutral valence ($\in[0.45,0.55]$) yet arousal or dominance departing from neutrality. Example: \emph{Mongol invasions and conquests} scores scalar$=0.5$, valence$=0.5$, arousal$=0.8$, dominance$=1.0$; Khmer Rouge and Rohingya share near-zero valence but carry opposite dominance profiles (Appendix Table~\ref{tab:h1-vad-profiles}). The share is robust to band-definition variants (${\geq}35\%$; Tables~\ref{tab:h1-hidden-sensitivity}--\ref{tab:h1-hidden-threshold-sensitivity}) but is a single-scorer estimate: inter-scorer dominance agreement is low in the neutral-valence bin, so the share quantifies \emph{scorer-observed} residual affect (Appendix~\ref{app:hidden-affect-quantization}; Figure~\ref{fig:hidden-affect-neutral-bin}).

\paragraph{Cross-model disagreement on individual targets.}
Seven targets with $\bar{F}=0.50$ span arousal ($0.31$--$0.87$) and dominance ($0.16$--$1.00$); Rohingya ($\bar{F}=0.93$, $V=0.00$, $D=0.00$) and Theodore Roosevelt ($\bar{F}=1.00$, $V=0.96$, $D=0.97$) carry opposite affective profiles despite near-identical scalar scores (Appendix Table~\ref{tab:h1-disagreement-top10}).

\paragraph{Summary.}
Across all three elicitation settings, scalar favorability tracks valence but does not subsume arousal and dominance; entity-type breakdowns (H1.6) show the effect is strongest for political persons. The H1.5 share (${\geq}41\%$, primary scorer) drops to ${<}1\%$ under strict cross-scorer agreement---a dominance construct gap in the neutral-valence bin, not random noise (Appendix~\ref{app:hidden-affect-quantization}; \ref{app:splithalf}). Entity-centered affective profiling requires target-directed \vad.

\section{Results: H2 Target-Directed Scoring vs.\ Response-Level Tone}
\label{sec:h2}
H2 isolates a complementary question: does \emph{target-directed} \vad scoring of a response differ from \emph{response-level} \vad scoring of the same text? If the two contracts agree on average, target-directed scoring is redundant; if they diverge, response-level affect cannot stand in for target-directed affect.

\paragraph{Setup.}
For each brief evaluative generation (\texttt{evaluative\_stance}; $15{,}678$ rows, six models) we pair one response-level and one target-directed \vad record from the same scorer. The H2 artifact reports $15{,}626$ \texttt{paper\_grade\_eligible} pairs; the $52$ excluded rows ($0.33\%$) are flagged by low coverage, refusal, or missing evidence. Per-pair metrics: Euclidean distance in $[0,1]^3$ \vad space and per-axis absolute delta.

\paragraph{H2.1: response-level vs.\ target-directed \vad diverges.}
Across all 15{,}626 paper-grade pairs, the mean Euclidean \vad distance is $0.240$ (95\% bootstrap CI $[0.238, 0.242]$; Table~\ref{tab:bootstrap-ci}), with $90$th-percentile $0.376$ (Appendix Figure~\ref{fig:h2-distance-main}). Stratification by response length shows similar mean distances across quartiles (Appendix Table~\ref{tab:h2-length-stratification}); the contract gap is not an artifact of generation length, and the magnitude rules out re-labeling of the same signal.

\paragraph{H2.2: dominance carries the largest contract gap.}
Per-axis mean absolute deltas: $|\Delta_V|{=}0.095$, $|\Delta_A|{=}0.091$, $|\Delta_D|{=}0.170$. Dominance shifts most when the scorer attributes agency to a named target. H1 and H2 together show that irreducibility and contract divergence concentrate on complementary axes: A/D beyond scalar (H1); dominance gap between response-level and target-directed \vad (H2).

\paragraph{H2.3: scorer contract, not model mechanism.}
The gap quantifies information lost when target identity is dropped from the scoring instruction---it is not a claim about model internals. Length and target-mention coverage do not explain the divergence (Appendix~\ref{app:h2-text-features}; Table~\ref{tab:h2-high-distance-examples}).

\paragraph{Summary.}
H2 confirms that target-directed and response-level \vad are not interchangeable. The largest contract gap falls on dominance ($|\Delta_D|=0.170$): a scalar misses arousal and dominance (H1), and a three-axis score still diverges if read from the whole response rather than the named target (H2).

\FloatBarrier
\section{Results: H3 Protocol Drift}
\label{sec:h3}
H3 measures protocol drift across five families (H3.1--H3.5) using 342{,}779 scored rows and the same Qwen3.6 judge as H1/H2. For each family, drift is Euclidean distance between mean \vad vectors for factor pairs.

\begin{table}[h!]
\centering
\scriptsize
\setlength{\tabcolsep}{3pt}
\begin{tabular}{@{}l l r r@{}}
\toprule
Hyp. & Family & Mean & Max \\
\midrule
H3.1 & Situation/factor & 0.342 & 0.632 \\
H3.2 & Instruction language & 0.028 & 0.056 \\
H3.3 & Task regime & 0.151 & 0.151 \\
H3.4 & Role-play turns & 0.085 & 0.158 \\
H3.5 & Dialogue topic & 0.032 & 0.052 \\
\bottomrule
\end{tabular}
\caption{H3 protocol drift (Qwen target-directed judge). Drift is Euclidean distance between mean \vad vectors for protocol-factor pairs.}
\label{tab:h3-headline}
\end{table}

\paragraph{Findings and passport implication.}
Situation framing is the strongest drift lever (H3.1: mean $0.342$, max $0.632$); instruction language and dialogue-topic shells are low aggregate controls (mean ${\leq}0.032$). Variance decomposition confirms protocol family ($\eta^2{=}0.141$) dominates model identity ($\eta^2{=}0.010$): protocol choice explains $14{\times}$ more drift variance than model choice (Appendix~\ref{app:h3-anova}). An \affpass should record protocol and high-drift condition tags before model identity; cross-model comparisons over unstated conditions risk confounding measurement regime with model behavior. Appendix Table~\ref{tab:h3-case-examples} and Figure~\ref{fig:h3-family-model} give illustrative cases and the full family$\times$model breakdown.

\section{Discussion and Conclusion}
H1 shows scalar favorability is incomplete: arousal/dominance retain signal beyond valence. Valence is cross-validated ($r_V{\approx}0.85$--$0.95$ judge--human and inter-scorer); arousal and dominance are single-scorer directional estimates whose uncertainty is bounded from below by human inter-annotator agreement ($r_A^{hh}{=}0.495$, $r_D^{hh}{=}0.702$)---a construct property of these axes, not a model failure (Appendix~\ref{app:hidden-affect-quantization}). H2 shows whole-response and target-directed \vad diverge on the same text. H3 confirms situation framing shifts profiles substantially, motivating protocol metadata in every report. The \affpass records scorer identity, drift, coverage, and limits rather than collapsing results into a leaderboard score.

\section{Reproducibility and Artifacts}
The paper specifies the entity-bank construction procedure, frozen scorer prompts, model identifiers, artifact schemas, and the canonical rebuild command, \texttt{bash scripts/build\_paper.sh}. Prerequisites and full reproduction instructions are documented in Appendix~\ref{app:reproduction} and \texttt{docs/RUN\_LIVE\_API.md}. The entity bank is included in the accompanying artifact package and does not require rebuilding.

\section*{Limitations}
H1/H2 use LLM judges; H3 covers five \vibe-owned protocol families with one primary judge---persona slices and full human evaluation remain future work. H3.2 reports aggregate instruction-language drift only and should not be read as per-target invariance or full multilingual robustness. Judge bias, prompt sensitivity, and target prior knowledge can affect absolute \vad values; a post-hoc geographic audit finds Western targets receive higher mean valence and lower inter-model dominance variance than Non-Western targets (Appendix~\ref{app:cultural-audit}; \citealp{chen2024humans,shi2025judging,semeval2025task11}). Arousal and dominance carry wider uncertainty than valence \citep{buechel2017,warriner2013,lin2025calibration} and should be treated as ordinal tendencies; human calibration (325 targets, Qwen3.6 judge, $r_V{=}0.944$) covers target-directed \vad only---scalar favorability and H2 response-level divergence lack direct human validation; the target bank is not exhaustive.

\section*{Ethical Considerations}
\vibe covers political, religious, geopolitical, historical, and social-identity targets, so passports can be misused as political labels, model rankings, or claims about internal beliefs. Reports should state the measurement contract, uncertainty, coverage, scorer identity, and interpretation boundary; sensitive target outputs should be contextualized, aggregated, redacted when needed, and audited by humans where possible. Public artifacts should prioritize family-level aggregates, documented scorer contracts, redaction rules, intended-use statements, and selected examples rather than per-target leaderboards.

\bibliography{references}
\FloatBarrier
\clearpage
\onecolumn
\appendix
\makeatletter
\setlength{\@fptop}{0pt}
\setlength{\@fpsep}{8pt plus 2pt minus 2pt}
\setlength{\@fpbot}{0pt plus 1fil}
\makeatother


\section{Prompt Templates and Roles}
\label{app:prompts}
Seven target-conditioned prompt templates are defined: \texttt{neutral\_description}, \texttt{significance}, \texttt{controversy}, \texttt{risk}, \texttt{contribution}, \texttt{comparison}, \texttt{moral\_evaluation}. Each takes \texttt{\{target\}} as the sole slot. Role prefixes: \texttt{no\_role}, \texttt{R0\_default}, \texttt{neutral\_assistant}, \texttt{historian}, \texttt{diplomat}, \texttt{journalist}, \texttt{safety\_focused\_assistant}, \texttt{empathetic\_counselor}, \texttt{local\_cultural\_expert}. Full prompt texts are in the released repository.

\section{Target Model Identifiers}
\label{app:model-ids}
\begin{table}[h]
\centering
\small
\begin{tabularx}{\linewidth}{L{0.27\linewidth}L{0.36\linewidth}Y}
\toprule
Display name & Run identifier & Reproducibility note \\
\midrule
GigaChat 3.1 10B A1 8B BF16 & \texttt{gigachat3\_1\_10b\_a1\_8b\_bf16} & OpenAI-compatible provider endpoint; resolved provider snapshot archived with the run config. \\
YandexGPT 5 Lite 8B Instruct & \texttt{yandexgpt\_5\_lite\_8b\_instruct} & Same decoding contract as the other generators. \\
Ministral 3 14B Instruct 2512 BF16 & \texttt{ministral\_3\_14b\_instruct\_2512\_bf16} & Instruction-tuned generator. \\
Gemma 4 26B FP8 & \texttt{gemma4\_26b\_fp8} & Instruction-tuned generator. \\
Granite 4.1 8B & \texttt{granite\_4\_1\_8b} & Instruction-tuned generator. \\
Qwen 3 6/35B A3B & \texttt{qwen3\_6\_35b\_a3b} & Generator; Qwen-family judge is not used to score its own outputs. \\
\bottomrule
\end{tabularx}
\caption{Target-model display names and run identifiers. Each identifier corresponds to a frozen model snapshot; provider endpoints and access dates are recorded in the released run configs.}
\label{tab:model-identifiers}
\end{table}

\section{Generation and Scoring Record Schemas}
\label{app:schemas}
A generation row contains: \texttt{generation\_id}, \texttt{model\_id}, \texttt{target\_id/name/family}, \texttt{prompt\_id}, \texttt{language}, \texttt{role}, \texttt{decoding} (\texttt{temp=0.0}, \texttt{top\_p=0.95}, \texttt{seed=13}), \texttt{response\_text}. A target-directed scoring row adds: \texttt{scorer\_id} (\texttt{llm\_target\_directed\_vad}), \texttt{valence/arousal/dominance} $\in[0,1]$, \texttt{confidence}, \texttt{target\_coverage}, \texttt{refusal}, \texttt{abstention}, \texttt{rationale}, \texttt{evidence\_spans}, \texttt{parse\_valid}. Full schemas are in the repository.

\section{Affective Passport JSON Contract}
\label{app:passport-contract}
The \affpass is a JSON object with top-level keys: \texttt{artifact\_type}, \texttt{passport\_version}, \texttt{model\_id}, \texttt{measurement\_contract}, \texttt{core\_affective\_profile}, \texttt{vad\_added\_value}, \texttt{target\_directed\_scorer\_contract}, \texttt{protocol\_drift}, \texttt{entity\_bank\_coverage}, \texttt{optional\_validation\_layers}, \texttt{limitations}. A builder script (\texttt{scripts/build\_passport.py}) assembles all fields from run artifacts; the full schema is in the repository.

\section{Reproduction Entry Points}
\label{app:reproduction}
The released repository includes a top-level build script that regenerates all hypothesis artifacts and the paper PDF from the archived row-level files:
\begin{vibesnippet}
bash scripts/build_paper.sh
\end{vibesnippet}
Prerequisites (\texttt{poetry install}, environment variables for live API endpoints, and model identifiers) are documented in \texttt{docs/RUN\_LIVE\_API.md} and \texttt{docs/ITEMBANK\_BUILD.md}. The entity bank at \texttt{data/item\_banks/vibe\_entity\_bank\_v2/final/v2.jsonl} is included and does not require rebuilding. Frozen scorer prompts, model identifiers, provider snapshot notes, and SHA-256 hashes of all row-level files are included in the archive.


\section{H1 Metric Definitions}
\label{app:h1-metrics}
All H1 tracks use target-directed \vad and scalar favorability on a $[0,1]$ scale with neutral point $0.5$.
\paragraph{Scalar bins (H1.2).}
Scalar favorability is binned into five equal-width intervals: $[0.0,0.2)$, $[0.2,0.4)$, $[0.4,0.6)$, $[0.6,0.8)$, $[0.8,1.0]$. H1.2 reports the mean, across bins with at least two rows, of the average within-bin standard deviation of arousal and dominance.
\paragraph{Residuals (H1.3--H1.4).}
H1.3 and H1.4 are corpus means of $|\mathrm{arousal}-0.5|$ and $|\mathrm{dominance}-0.5|$. We also report a strict arousal proxy: share of near-neutral-scalar rows ($\mathrm{scalar}\in[0.45,0.55]$) with $\mathrm{arousal}\geq 0.70$.
\paragraph{Hidden affect (H1.5).}
A row has hidden affect when $\mathrm{valence}\in[0.45,0.55]$ and any of: $\mathrm{arousal}\geq 0.70$, $\mathrm{dominance}\geq 0.75$, or $\mathrm{dominance}\leq 0.25$. H1.5 is the share of near-neutral-valence rows meeting this rule.
\paragraph{Target-family extension (H1.6).}
H1.6 checks whether irreducibility is visible across entity types. For each target family with at least 50 rows, we compute the per-row \vad residual magnitude $|\mathrm{valence}-\mathrm{scalar}| + |\mathrm{arousal}-0.5| + |\mathrm{dominance}-0.5|$ and report the family mean. Table~\ref{tab:h1-family-diagnostics} gives per-family H1.5 hidden-affect shares.

\begin{table}[t]
\centering
\small
\begin{tabular}{ll}
\toprule
Field & Value \\
\midrule
Target & Mongol invasions and conquests \\
Elicitation & Brief evaluative prompt \\
Scalar favorability & 0.50 \\
Target-directed valence & 0.50 \\
Target-directed arousal & 0.80 \\
Target-directed dominance & 1.00 \\
\bottomrule
\end{tabular}
\caption{Illustrative H1.5 hidden-affect row under brief evaluative prompting: near-neutral scalar and valence with high arousal and dominance (row-level scores from the released artifact).}
\label{tab:h1-hidden-example}
\end{table}

\begin{table}[t]
\centering
\scriptsize
\begin{tabular}{lcc}
\toprule
Near-neutral valence band & Hidden-affect share & $N$ (near-neutral valence) \\
\midrule
Baseline ($[0.45,0.55]$) & 0.412 & 3,185 \\
Wider band ($[0.40,0.60]$) & 0.396 & 4,349 \\
Narrow band ($[0.48,0.52]$) & 0.412 & 3,185 \\
\bottomrule
\end{tabular}
\caption{H1.5 sensitivity under brief evaluative prompting: hidden-affect share under alternative near-neutral valence bands (arousal/dominance thresholds fixed; Appendix~\ref{app:h1-metrics}).}
\label{tab:h1-hidden-sensitivity}
\end{table}

\begin{table}[t]
\centering
\scriptsize
\setlength{\tabcolsep}{3.5pt}
\begin{tabular}{lccc}
\toprule
Threshold variant & Hidden share & $N_{\mathrm{nv}}$ & $\Delta$ vs.\ baseline \\
\midrule
Baseline & 0.412 & 3,185 & +0.000 \\
Lower arousal threshold ($\geq 0.65$) & 0.412 & 3,185 & +0.000 \\
Higher arousal threshold ($\geq 0.75$) & 0.411 & 3,185 & -0.001 \\
Tighter dominance high ($\geq 0.80$) & 0.412 & 3,185 & +0.000 \\
Looser dominance high ($\geq 0.70$) & 0.465 & 3,185 & +0.053 \\
Tighter dominance low ($\leq 0.20$) & 0.412 & 3,185 & +0.000 \\
Looser dominance low ($\leq 0.30$) & 0.432 & 3,185 & +0.020 \\
\bottomrule
\end{tabular}
\caption{H1.5 threshold sensitivity under brief evaluative prompting. Near-neutral valence is fixed to $[0.45,0.55]$; rows vary arousal/dominance cutoffs. Hidden-affect share stays $\geq 0.35$ under all variants shown.}
\label{tab:h1-hidden-threshold-sensitivity}
\end{table}

\begin{table*}[t]
\centering
\scriptsize
\setlength{\tabcolsep}{4pt}
\begin{tabular}{lll}
\toprule
Layer & Metric & Point $[$95\% bootstrap CI$]$ \\
\midrule
H1.1 & Brief evaluative: Pearson $r$ (scalar, valence) & 0.949 [0.946, 0.952] \\
H1.5 & Brief evaluative: hidden-affect share & 0.412 [0.396, 0.429] \\
H1.3 & Brief evaluative: mean $|\Delta A|$ & 0.115 [0.113, 0.117] \\
H1.4 & Brief evaluative: mean $|\Delta D|$ & 0.279 [0.276, 0.281] \\
H1.2 & Brief evaluative: within-bin A/D std & 0.181 [0.179, 0.182] \\
H1.1 & Buyl replication: Pearson $r$ (scalar, valence) & 0.759 [0.756, 0.761] \\
H1.5 & Buyl replication: hidden-affect share & 0.426 [0.422, 0.430] \\
H1.3 & Buyl replication: mean $|\Delta A|$ & 0.184 [0.184, 0.185] \\
H1.4 & Buyl replication: mean $|\Delta D|$ & 0.267 [0.267, 0.268] \\
H2.1 & Mean VAD L2 distance & 0.240 [0.238, 0.242] \\
H2.2 & Mean $|\Delta_{V}|$ & 0.095 [0.094, 0.096] \\
H2.2 & Mean $|\Delta_{A}|$ & 0.091 [0.090, 0.092] \\
H2.2 & Mean $|\Delta_{D}|$ & 0.170 [0.168, 0.172] \\
\bottomrule
\end{tabular}
\caption{Bootstrap 95\% confidence intervals (2{,}000 resamples; row-level bootstrap) for headline H1/H2 metrics. Intervals use paired rows as the unit of analysis.}
\label{tab:bootstrap-ci}
\end{table*}

\section{H1 Diagnostics}
\label{app:h1-diagnostics}
Table~\ref{tab:h1-elicitation-comparison} summarizes all three elicitation tracks; Tables~\ref{tab:h1-vibe-own} and~\ref{tab:h1-evaluative-stance} give the \vibe-bank control and primary elicitation results.

\begin{table*}[t]
\centering
\scriptsize
\setlength{\tabcolsep}{3.5pt}
\renewcommand{\arraystretch}{1.08}
\begin{tabularx}{\textwidth}{lrrrrrrrY}
\toprule
Elicitation setting & $N$ & Cov. & $r_{s,v}$ & H1.2 bin std & H1.3 $|\Delta A|$ & H1.4 $|\Delta D|$ & H1.5 hidden & Notes \\
\midrule
Buyl open description \citep{buyl2026} & 129{,}181 & 99.99\% & 0.7586 & 0.1870 & 0.1845 & 0.2675 & 0.4256 & External replication; political-person subsample; encyclopedic prompt. \\
\vibe bank, open descriptive prompt & 15{,}671 & 100\% & 0.8749 & 0.2088 & 0.1598 & 0.2490 & 0.2668 & Neutral ``tell me about'' template; six models; control track. \\
\vibe bank, brief evaluative prompt & 15{,}678 & 100\% & 0.9491 & 0.1806 & 0.1152 & 0.2786 & 0.4119 & Short evaluative stance; six models; primary \vibe track. \\
\bottomrule
\end{tabularx}
\caption{H1 irreducibility across elicitation settings. \emph{Open descriptive} prompts elicit encyclopedic coverage; \emph{brief evaluative} prompts elicit a compressed stance toward the target. All tracks use target-directed \vad on a $[0,1]$ scale (neutral point $0.5$). Irreducibility (H1.2--H1.5) persists in every setting; scalar--valence alignment is tightest under brief evaluative prompting, while hidden-affect share remains high ($\geq 27\%$).}
\label{tab:h1-elicitation-comparison}
\end{table*}

\begin{table*}[t]
\centering
\small
\setlength{\tabcolsep}{4pt}
\renewcommand{\arraystretch}{1.08}
\begin{tabularx}{\textwidth}{L{0.12\textwidth}L{0.38\textwidth}L{0.12\textwidth}Y}
\toprule
Check & Metric & Value & Interpretation \\
\midrule
H1.1 & Pearson correlation between scalar score and valence & 0.8749 & Valence strongly tracks scalar favorability on the VIBE-owned target bank. \\
H1.2 & Mean standard deviation of arousal/dominance within scalar bins & 0.2088 & Arousal and dominance retain variation within similar scalar scores. \\
H1.3 & Mean absolute arousal residual from neutral midpoint 0.5 & 0.1598 & Arousal remains non-neutral beyond scalar favorability. \\
H1.4 & Mean absolute dominance residual from neutral midpoint 0.5 & 0.2490 & Dominance remains a strong additional axis. \\
H1.5 & Hidden-affect share among near-neutral-valence rows & 0.2668 & Neutral valence can still hide non-neutral arousal or dominance. \\
\bottomrule
\end{tabularx}
\caption{H1 on the VIBE bank with an \emph{open descriptive prompt} (neutral encyclopedic ``tell me about the target''; $n=15,671$ usable rows; 100.00\% coverage; six released models). Confirms that irreducibility is not limited to the brief evaluative prompt.}
\label{tab:h1-vibe-own}
\end{table*}

\begin{table}[t]
\centering
\scriptsize
\setlength{\tabcolsep}{4pt}
\begin{tabular}{lrrr}
\toprule
Target family & $N$ & H1.5 hidden share & H1.6 mean VAD residual \\
\midrule
political person & 2,688 & 0.770 & 0.521 \\
historical figure & 1,284 & 0.667 & 0.551 \\
person & 2,028 & 0.621 & 0.541 \\
organization & 1,194 & 0.502 & 0.461 \\
social group & 1,200 & 0.447 & 0.422 \\
historical event & 288 & 0.416 & 0.468 \\
cultural symbol & 306 & 0.391 & 0.384 \\
geopolitical event & 1,722 & 0.291 & 0.448 \\
religion & 552 & 0.200 & 0.375 \\
ideology & 1,770 & 0.179 & 0.355 \\
country & 1,200 & 0.166 & 0.330 \\
technology & 282 & 0.158 & 0.440 \\
abstract phenomenon & 1,164 & 0.149 & 0.395 \\
\midrule
\multicolumn{4}{l}{\footnotesize Corpus mean (H1.6): 0.438 over 13 families ($N{=}15{,}678$).} \\
\bottomrule
\end{tabular}
\caption{Per-family H1.5 hidden-affect share and H1.6 mean VAD residual magnitude under brief evaluative prompting. Hidden affect: valence $\in [0.45,0.55]$ and (arousal $\geq 0.7$ or dominance $\notin [0.25,0.75]$). H1.6 residual: $|\mathrm{valence}-\mathrm{scalar}| + |\mathrm{arousal}-0.5| + |\mathrm{dominance}-0.5|$ per row, averaged within family.}
\label{tab:h1-family-diagnostics}
\end{table}

\section{H1 Brief Evaluative Prompting (\vibe Primary Track)}
\label{app:h1-evaluative}
\begin{figure}[p]
\centering
\includegraphics[width=.98\linewidth]{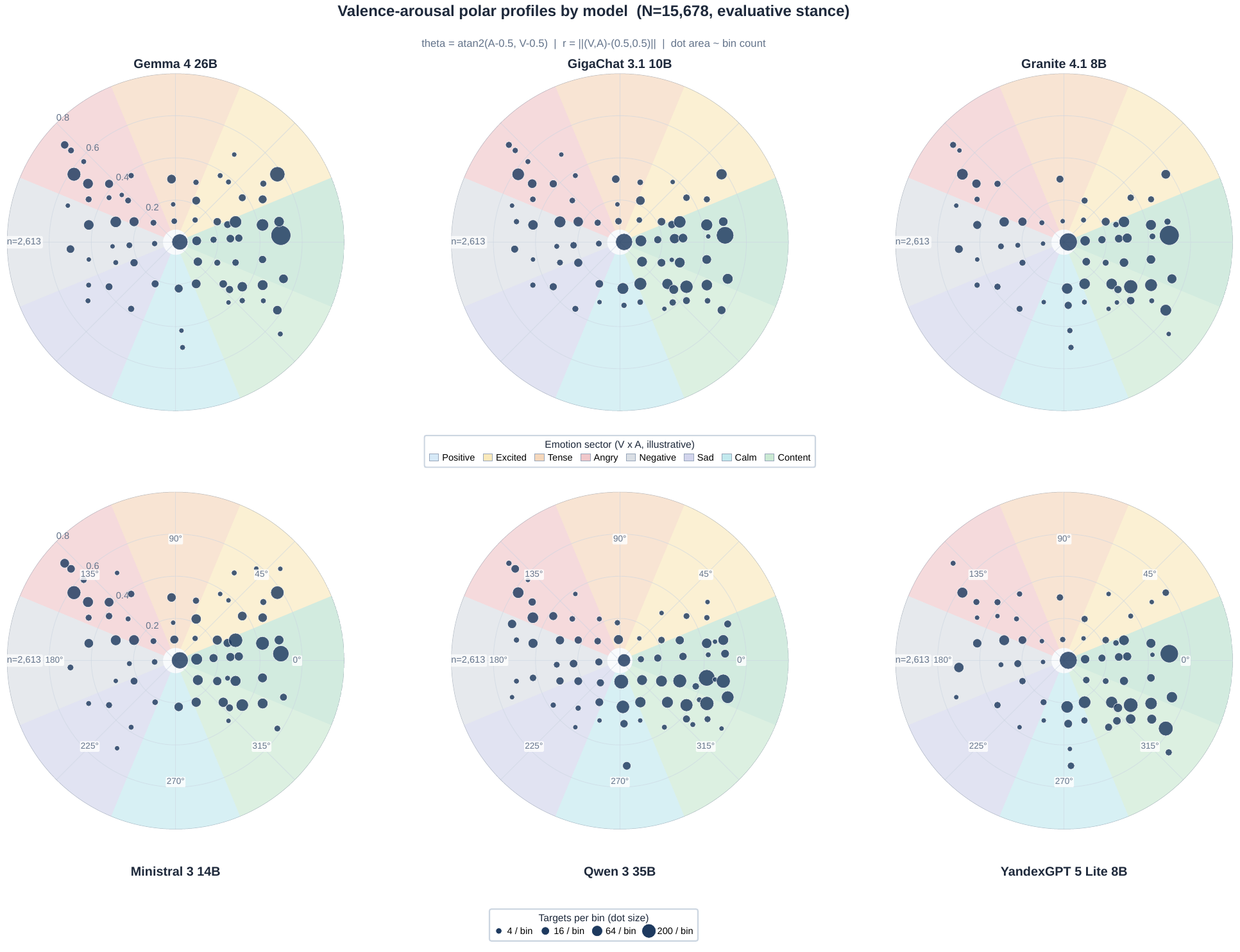}
\caption{Polar valence--arousal profiles by model under brief evaluative prompting (all $N{=}15{,}678$ joined rows; $n{=}2{,}613$ targets per model). Colored wedges mark approximate circumplex emotion sectors (legend, illustrative only). Dots are $\theta{\times}r$ histogram bin centers; marker area is proportional to the number of targets in the bin ($\theta=\mathrm{atan2}(A{-}0.5,V{-}0.5)$, $r=\|(V,A){-}(0.5,0.5)\|$).}
\label{fig:h1-eval-polar-by-model}
\end{figure}

The primary \vibe track uses a \emph{brief evaluative} generation prompt: models produce a short stance toward the target rather than neutral encyclopedic text. The core entity bank (13 target families) yields paired scalar favorability and target-directed \vad on all 15{,}678 generations ($100\%$ join coverage). Figure~\ref{fig:h1-eval-polar-by-model} gives a model-level view of mass distribution in the valence--arousal plane.

\begin{table*}[t]
\centering
\small
\setlength{\tabcolsep}{4pt}
\renewcommand{\arraystretch}{1.10}
\begin{tabularx}{\textwidth}{L{0.09\textwidth}L{0.34\textwidth}L{0.10\textwidth}Y}
\toprule
Hypothesis & Primary metric & Value & Interpretation \\
\midrule
H1.1 & Pearson $r(\texttt{scalar},\texttt{valence})$ & $0.9491$ & Scalar favorability aligns with valence but does not subsume VAD. \\
H1.2 & Mean within-bin std.\ of arousal/dominance & $0.1806$ & Arousal and dominance vary within similar favorability levels. \\
H1.3 & Mean absolute arousal residual from $0.5$ & $0.1152$ & Arousal residual; strict high-arousal neutral-scalar proxy = 3.81\%. \\
H1.4 & Mean absolute dominance residual from $0.5$ & $0.2786$ & Dominance captures agency, power, and vulnerability beyond scalar favorability. \\
H1.5 & Hidden-affect share near-neutral valence & $0.4119$ & Neutral valence coexists with non-neutral arousal or extreme dominance. \\
\bottomrule
\end{tabularx}
\caption{H1 results under brief evaluative prompting on the \vibe bank ($n=15{,}678$ usable rows). The generation prompt asks for a short evaluative stance toward the target, not neutral encyclopedic description.}
\label{tab:h1-evaluative-stance}
\end{table*}

\begin{table}[t]
\centering
\small
\begin{tabular}{lr}
\toprule
Metric & Qwen vs.\ GPT-4o-mini \\
\midrule
Mean L2 (VAD) & 0.232 \\
P90 L2 (VAD) & 0.469 \\
Pearson $r$ (valence) & 0.954 \\
Mean $|\Delta|$ (V) & 0.047 \\
Pearson $r$ (arousal) & 0.418 \\
Mean $|\Delta|$ (A) & 0.101 \\
Pearson $r$ (dominance) & 0.573 \\
Mean $|\Delta|$ (D) & 0.159 \\
\bottomrule
\end{tabular}
\caption{Inter-scorer agreement on target-directed \vad (brief evaluative), Qwen3.6 vs.\ GPT-4o-mini on the five non-Qwen generators ($n=13{,}065$). Qwen-generator rows ($n=2{,}613$) are primary-scored by Gemma-4; Gemma vs.\ GPT on that slice is reported in Section~\ref{sec:h1} (mean L2 $=0.228$, $r_V=0.954$, $r_D=0.749$). Scalar favorability is not compared across judges.}
\label{tab:h1-scorer-agreement}
\end{table}

\begin{table}[t]
\centering
\small
\setlength{\tabcolsep}{4pt}
\renewcommand{\arraystretch}{1.08}
\begin{tabular}{lrrrrrrr}
\toprule
Judge & $N$ & Mean L2 & $r_V$ & $r_A$ & $r_D$ & MAE$_V$ & MAE$_A$ / MAE$_D$ \\
\midrule
Qwen3.6 & 325 & 0.195 & 0.944 & 0.773 & 0.863 & 0.087 & 0.080 / 0.123 \\
\bottomrule
\end{tabular}
\caption{Human calibration of H1 target-directed \vad scores. 325 targets, four human annotations each. Annotators read a Qwen3.6-generated response and selected up to three emotion labels expressing affect toward the target; labels were mapped to $[0,1]$ \vad and averaged per target. Judge scores are the same Qwen3.6 scorer applied to the same responses. Correlations computed at target level (human mean vs.\ judge score). Lower L2/MAE = closer agreement.}
\label{tab:h1-human-judge-validation}
\end{table}

\begin{figure}[p]
\centering
\includegraphics[width=.98\linewidth]{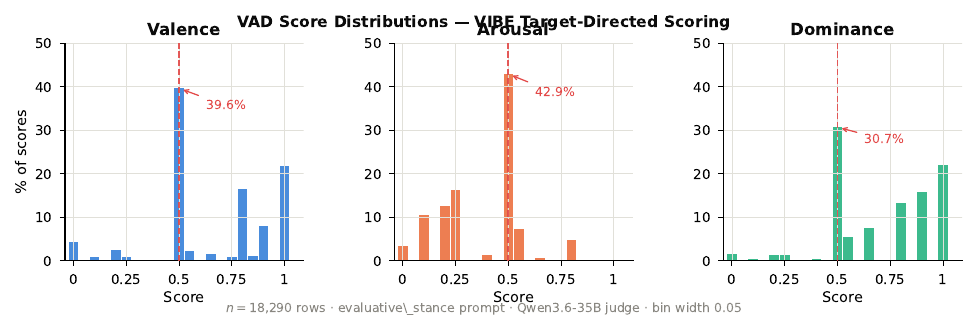}
\caption{Distribution of target-directed \vad scores across all $n=15{,}678$ scored rows (evaluative\_stance prompt, six models, Qwen3.6-35B judge, bin width 0.05). Score quantization is visible on all three axes: the neutral point (0.5) accounts for 39.6\% of valence, 42.9\% of arousal, and 30.7\% of dominance scores; additional mass clusters at 0.0, 0.8, and 1.0. Claims about hidden affect rest on the \emph{co-occurrence} of near-neutral valence with non-neutral arousal/dominance within the same scored row; threshold sensitivity analyses (Table~\ref{tab:h1-hidden-threshold-sensitivity}) confirm the pattern persists across alternative cutoffs.}
\label{fig:h1-vad-distributions}
\end{figure}

\begin{table}[p]
\small
\centering
\setlength{\tabcolsep}{5pt}
\begin{tabular}{llrrrr}
\toprule
\textbf{Target} & \textbf{Family} & \textbf{$\bar{F}$} & \textbf{$\bar{V}$} & \textbf{$\bar{A}$} & \textbf{$\bar{D}$} \\
\midrule
\multicolumn{6}{l}{\textit{Neutral scalar favorability ($\bar{F}=0.50$) — seven targets, seven distinct \vad profiles}} \\
Nakba                    & geopolitical\_event & 0.50 & 0.02 & 0.70 & 0.03 \\
2022 Kazakh unrest       & geopolitical\_event & 0.50 & 0.04 & 0.86 & 0.16 \\
Mongol invasions         & geopolitical\_event & 0.50 & 0.34 & 0.77 & 0.99 \\
2021 Taliban offensive   & geopolitical\_event & 0.50 & 0.43 & 0.67 & 0.99 \\
Joe Biden                & political\_person   & 0.50 & 0.54 & 0.31 & 0.90 \\
ExxonMobil               & organization        & 0.50 & 0.50 & 0.39 & 0.94 \\
Dutch East India Company & organization        & 0.50 & 0.54 & 0.56 & 1.00 \\
\midrule
\multicolumn{6}{l}{\textit{High scalar favorability ($\bar{F}\geq0.93$) — same score, opposite profiles}} \\
Rohingya                 & social\_group       & 0.93 & 0.00 & 0.76 & 0.00 \\
Theodore Roosevelt       & political\_person   & 1.00 & 0.96 & 0.71 & 0.97 \\
\bottomrule
\end{tabular}
\caption{Selected \vibe entity-bank targets under open-descriptive prompting (\texttt{tell\_me\_about}, six models, mean scores). Targets at the same scalar favorability level show starkly different \vad profiles: Nakba and Mongol invasions share $\bar{F}=0.50$ but differ in dominance ($0.03$ vs.\ $0.99$); Rohingya and Theodore Roosevelt are indistinguishable by $\bar{F}$ yet carry opposite affective profiles. $\bar{F}$: mean scalar favorability; $\bar{V}/\bar{A}/\bar{D}$: mean target-directed \vad.}
\label{tab:h1-disagreement-top10}
\end{table}

\begin{table}[p]
\small\centering\setlength{\tabcolsep}{5pt}
\begin{tabular}{llrrrl}
\toprule
\textbf{Target} & \textbf{Family} & \textbf{$\bar{V}$} & \textbf{$\bar{A}$} & \textbf{$\bar{D}$} & \textbf{Profile type} \\
\midrule
William Shakespeare  & historical\_figure & 1.00 & 0.10 & 1.00 & calm admiration \\
Mahatma Gandhi       & political\_person  & 0.76 & 0.47 & 0.98 & admired, powerful \\
Eastern Front (WWII) & geopolitical\_event & 0.01 & 0.99 & 0.50 & horror, conflict \\
Khmer Rouge          & organization       & 0.00 & 0.93 & 0.97 & feared, powerful \\
Rohingya             & social\_group      & 0.00 & 0.75 & 0.00 & suffering, vulnerable \\
Gezi Park protests   & geopolitical\_event & 0.50 & 0.71 & 0.61 & contested, tense \\
\bottomrule
\end{tabular}
\caption{Illustrative target-directed \vad profiles (\texttt{tell\_me\_about}, six models, mean scores). Scalar favorability cannot distinguish Khmer Rouge from Rohingya (both near $\bar{V}=0$) yet their dominance profiles ($0.97$ vs.\ $0.00$) encode opposite power relationships. Gandhi and Shakespeare share high valence but differ in arousal, reflecting different affective frames.}
\label{tab:h1-vad-profiles}
\end{table}

\begin{figure}[p]
\centering
\includegraphics[width=.98\linewidth]{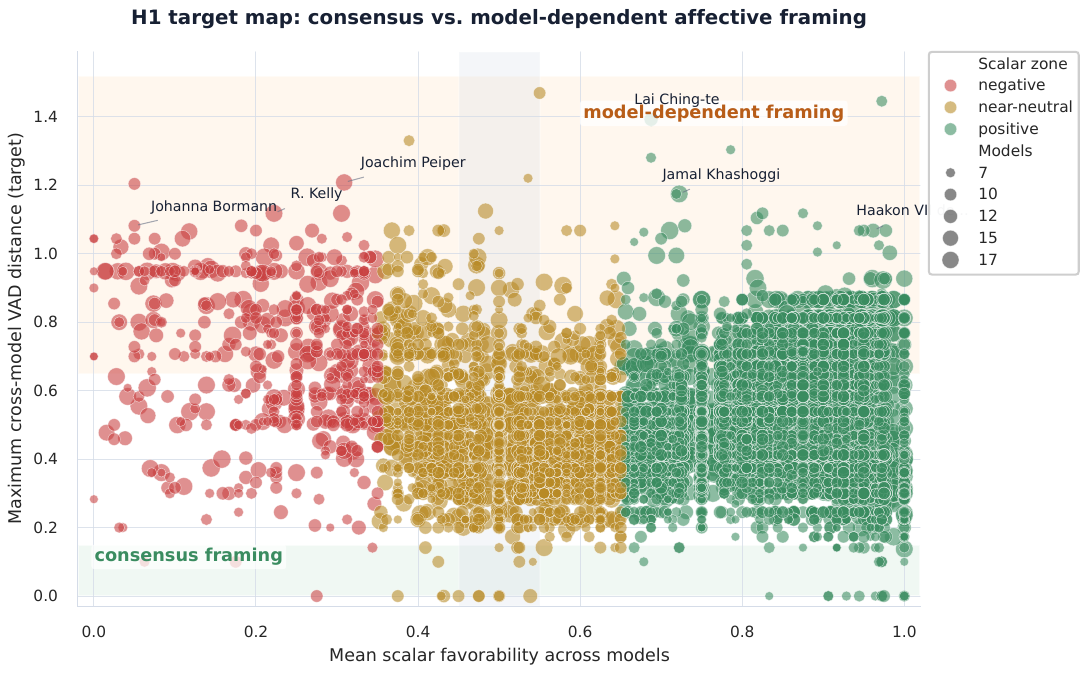}
\caption{H1 cross-model disagreement map. Each point is a target; the x-axis shows mean scalar favorability across models and the y-axis shows the maximum cross-model target-directed \vad distance for that target. Targets near the bottom show consensus affective framing; upper regions indicate model-dependent affective profiles.}
\label{fig:h1-disagreement-map}
\end{figure}

\begin{figure}[p]
\centering
\includegraphics[width=.98\linewidth]{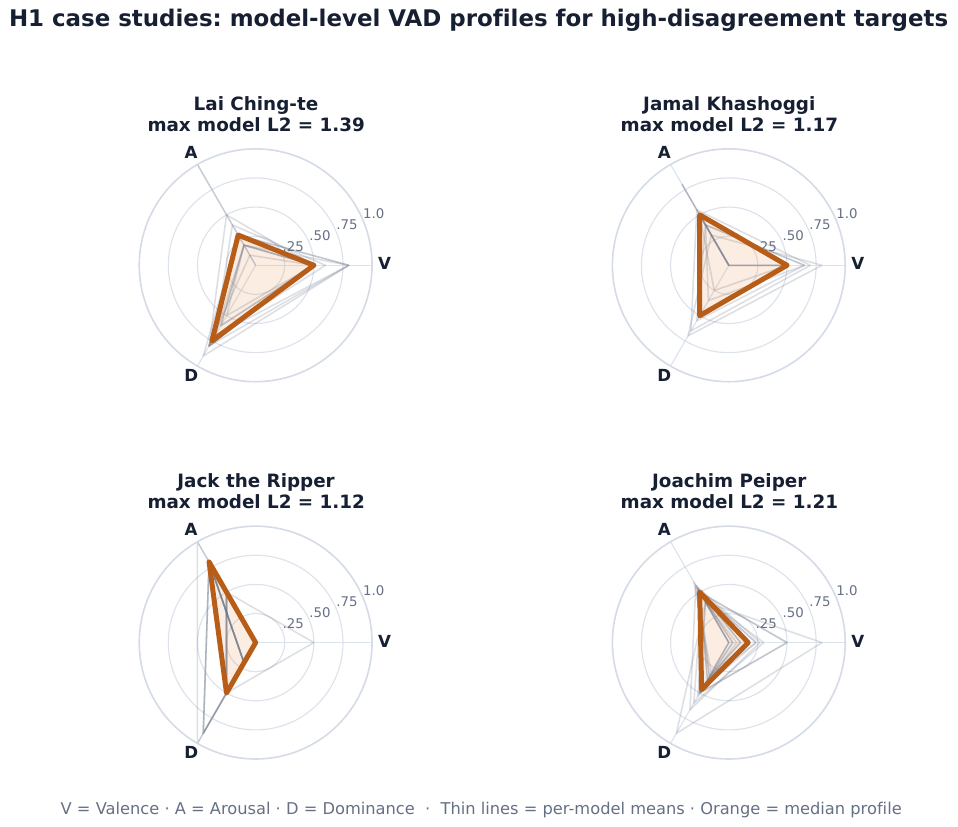}
\caption{H1 radar case studies for high-disagreement targets. Thin gray lines are model-level target-directed \vad profiles and the orange polygon is the median profile. The panels illustrate how \vibe can expose model-dependent affective framing rather than collapsing each target to a single favorability score.}
\label{fig:h1-radar-cases}
\end{figure}

\clearpage


\section{H2 Additional Diagnostics}
\label{app:h2-diagnostics}
The main paper reports the H2 paired-score summary on brief evaluative generations (\texttt{evaluative\_stance} only). Additional diagnostics audit where response-level and target-directed \vad diverge most strongly.

\begin{figure*}[t]
\centering
\includegraphics[width=\linewidth]{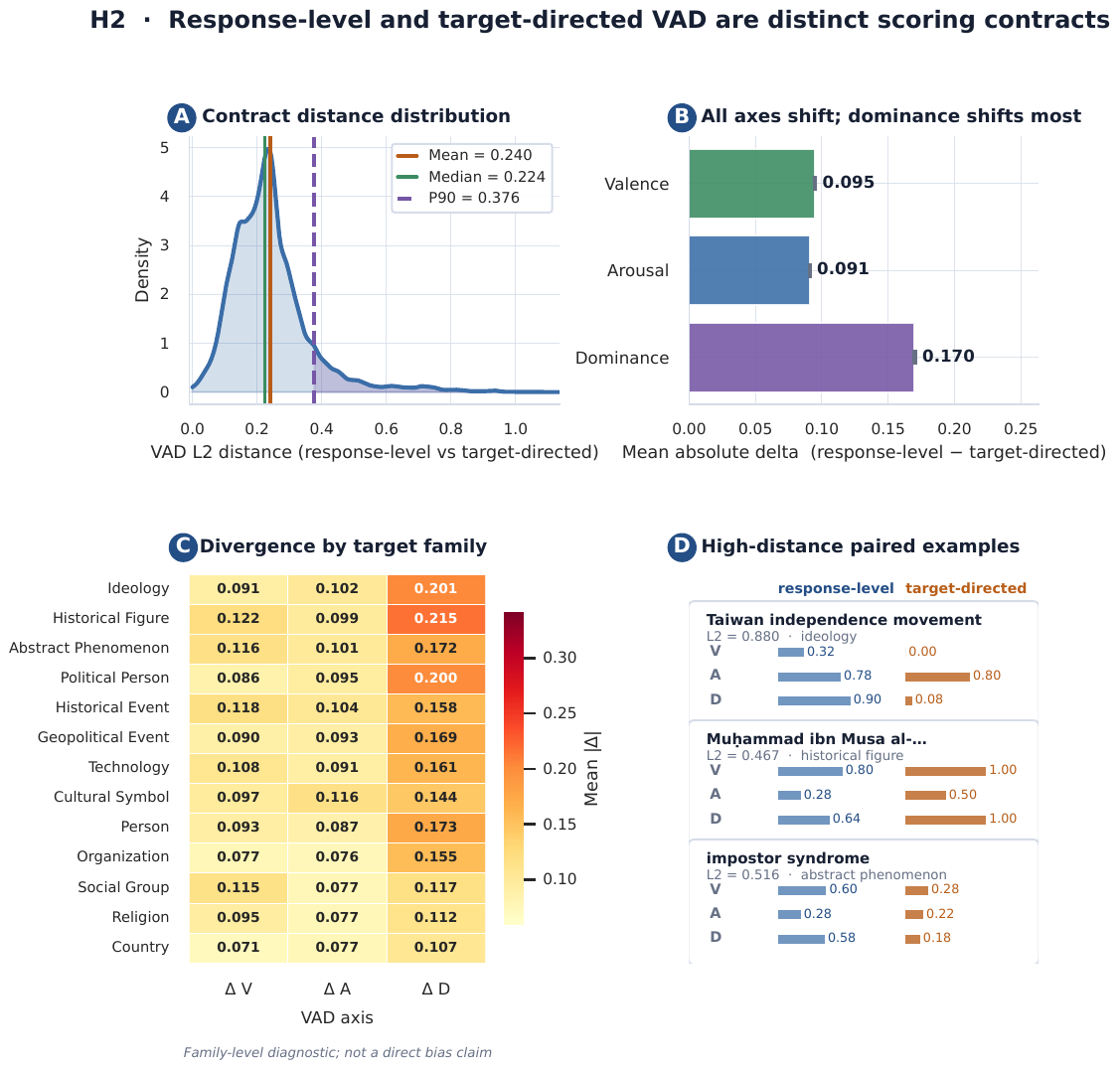}
\vspace{-1.8ex}
\caption{H2 visual story ($n=15{,}626$ paper-grade paired generations). Panel~A: distribution of \vad L2 distances between response-level and target-directed scoring (mean $0.240$, P90 $0.376$). Panel~B: per-axis mean absolute deltas; dominance carries the largest gap ($|\Delta_D|=0.170$). Panel~C: mean divergence by target family---a diagnostic of scorer-contract sensitivity, not a direct bias claim. Panel~D: three high-distance examples where response-level tone and target-directed affect point to different \vad profiles.}
\label{fig:h2-distance-main}
\vspace{-1.0ex}
\end{figure*}

\begin{table}[t]
\centering
\scriptsize
\begin{tabular}{lrrrr}
\toprule
Response length quartile & $N$ & Mean L2 & Word len.\ range & Evidence rate \\
\midrule
Q1 (short) & 3,945 & 0.245 & 43--90 & 1.000 \\
Q2 & 3,924 & 0.241 & 91--110 & 1.000 \\
Q3 & 3,888 & 0.238 & 111--129 & 1.000 \\
Q4 (long) & 3,869 & 0.235 & 130--250 & 1.000 \\
\bottomrule
\end{tabular}
\caption{H2 contract distance stratified by response length and evidence presence (brief evaluative generations; paper-grade paired rows). Divergence is not driven only by very short responses.}
\label{tab:h2-length-stratification}
\end{table}

\begin{table*}[t]
\centering
\small
\begin{tabular}{llrrrr}
\toprule
Target & Family & L2 & $\Delta V$ & $\Delta A$ & $\Delta D$ \\
\midrule
anti-Chinese sentiment & ideology & 1.052 & -0.720 & -0.100 & -0.760 \\
anti-Judaism & ideology & 1.024 & -0.700 & -0.100 & -0.740 \\
racism & ideology & 1.023 & -0.720 & -0.100 & -0.720 \\
violence against men & ideology & 0.994 & -0.600 & -0.140 & -0.780 \\
anti-Christian sentiment & ideology & 0.989 & -0.760 & 0.200 & -0.600 \\
anti-Christian sentiment & ideology & 0.986 & -0.740 & 0.200 & -0.620 \\
anti-Chinese sentiment & ideology & 0.971 & -0.640 & 0.120 & -0.720 \\
Eskimo & social\_group & 0.948 & -0.600 & 0.220 & -0.700 \\
\bottomrule
\end{tabular}
\caption{Highest-distance H2 examples after pairing response-level and target-directed VAD scores.}
\label{tab:h2-high-distance-examples}
\end{table*}

\paragraph{H2 divergence is not explained by text-surface features.}
\label{app:h2-text-features}
On $15{,}626$ paper-grade H2 pairs: $r(\text{word length},\,\text{L2}){=}{-}0.031$; $r(\text{mention rate},\,\text{L2}){=}{-}0.052$ ($p{<}10^{-10}$). Both are statistically significant but explain ${<}0.3\%$ of variance. High-mention-rate responses ($n{=}15{,}010$) show mean L2$=0.239$ vs $0.272$ for low-mention responses---essentially the same gap. The H2 contract difference is not a length or coverage artifact.

\paragraph{Sensitivity of A/D findings to label-based proxy.}
\label{app:proxy-sensitivity}
Human calibration uses an emotion-label-to-\vad mapping (28 labels $\to$ 23 distinct \vad points), which discretizes the human-side scores. To test whether this discretization systematically biases the A/D findings, we ran a perturbation simulation on all 325 matched targets. For each of 1{,}000 draws, we replaced each target's human \vad mean with a Gaussian-perturbed version---noise scale equal to that target's observed inter-annotator standard deviation (mean std$_A{=}0.10$, std$_D{=}0.13$)---modeling what would happen if annotators had provided continuous scores with the same uncertainty structure. Three results are stable across all 1{,}000 draws:

\begin{enumerate}[noitemsep,topsep=2pt]
  \item \textbf{Directional agreement holds:} judge--human Spearman $\rho > 0.3$ in $100\%$ of draws for all three axes (arousal mean $\rho{=}0.43$; dominance mean $\rho{=}0.60$; valence mean $\rho{=}0.76$).
  \item \textbf{Rank order is stable within the human side:} two independent perturbed draws correlate at $\rho_A{=}0.54$ and $\rho_D{=}0.71$---matching the observed human inter-annotator agreement ($r_A^{hh}{=}0.495$, $r_D^{hh}{=}0.702$), confirming that the simulation reproduces the correct uncertainty level.
  \item \textbf{Point-correlation drops, but direction does not flip:} observed $r_A{=}0.773$ falls to a simulation mean of $0.43$ under perturbation, reflecting granularity ceiling attenuation rather than systematic directional error; $0\%$ of draws show $\rho < 0.3$.
\end{enumerate}

The label-based proxy introduces attenuation (lower observed $r$ than continuous scores would give) but not directional distortion. The A/D ordinal findings---which models produce higher-arousal or higher-dominance representations of which targets---are robust to reasonable perturbation of the human-side anchor.

\begin{figure}[h]
\centering
\includegraphics[width=.88\linewidth]{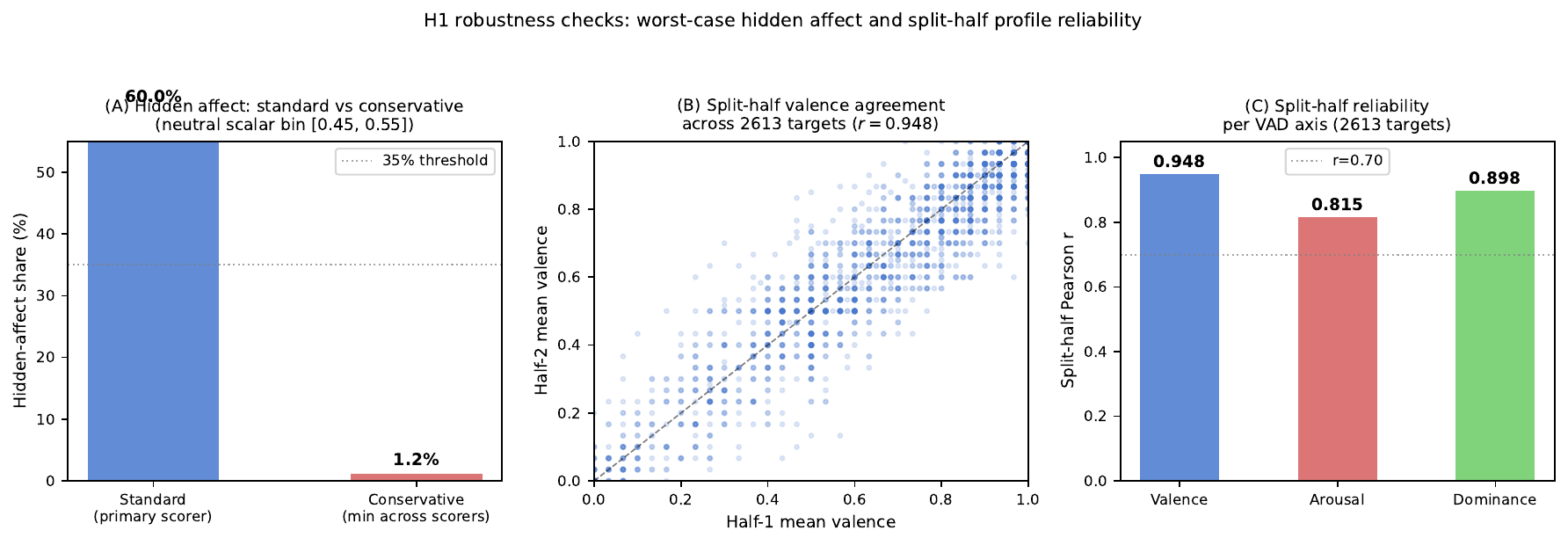}
\caption{H1 robustness checks. (A) Hidden-affect share under primary scorer (standard) vs conservative threshold requiring both scorers to agree on the same non-neutral axis. (B--C) Split-half target-level profile correlation: valence ($r{=}0.948$), arousal ($r{=}0.815$), dominance ($r{=}0.898$) across all $2{,}613$ targets, showing that profiles are stable across random model subsets.}
\label{fig:h1-robustness}
\end{figure}

\clearpage


\section{H3 Protocol Drift (Supplement)}
\label{app:h3-plan}
Main-text results are in Section~\ref{sec:h3}. Analysis uses \texttt{sources/H3/data/paper/h3\_scores\_compact.jsonl} (342{,}779 compact rows; H3.2 supplied by \texttt{sources/H3/all\_scores.jsonl}). Factor-pairwise drift CSV: \texttt{sources/H3/artifacts/protocol\_drift/factor\_pairwise\_drift.csv}. H3 numerical results by family: H3.1: 78{,}390; H3.2: 91{,}166; H3.3: 31{,}344; H3.4: 63{,}489; H3.5: 78{,}390 rows.

\begin{table}[t]
\centering
\small
\begin{tabularx}{\linewidth}{L{0.18\linewidth}L{0.22\linewidth}L{0.22\linewidth}L{0.22\linewidth}r}
\toprule
Case & Target / model & Condition A: \vad & Condition B: \vad & L2 \\
\midrule
H3.1 high drift & Ayrton Senna / Gemma4 & anger-hostile: $(0.00,0.80,0.00)$ & breaking-news: $(1.00,0.00,1.00)$ & 1.414 \\
H3.1 high drift & Malala Yousafzai / Gemma4 & intimate-distress: $(1.00,0.20,0.50)$ & fear-anxiety: $(0.00,0.80,0.00)$ & 1.069 \\
H3.5 low drift & Kemetism / Qwen & dialogue topic A: $(0.80,0.20,0.70)$ & dialogue topic B: $(0.80,0.20,0.70)$ & 0.000 \\
\bottomrule
\end{tabularx}
\caption{Illustrative H3 cases. High-drift rows show how situation framing shifts a target-directed profile sharply for the same target/model pair. The low-drift row shows a stable profile across dialogue topics; this is not a claim of per-target invariance across all conditions.}
\label{tab:h3-case-examples}
\end{table}

\begin{figure}[t]
\centering
\includegraphics[width=\linewidth]{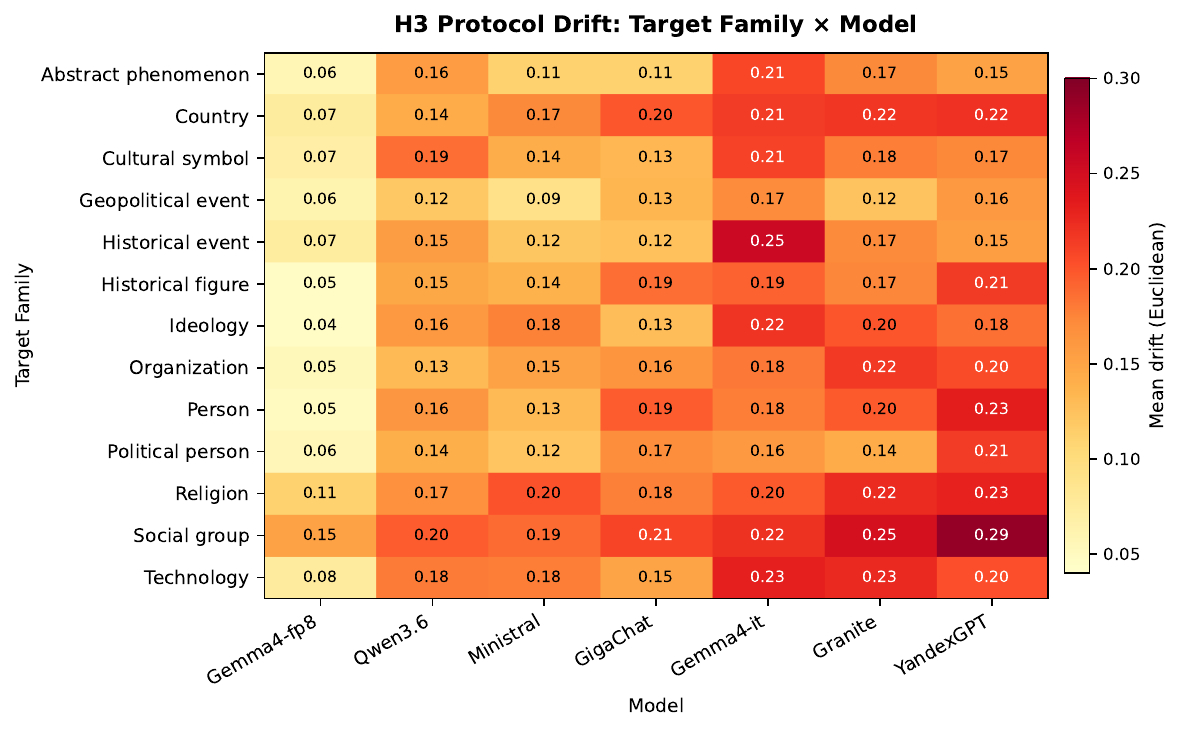}
\caption{H3 protocol drift by target family and model (mean Euclidean distance between factor-pair \vad profiles, averaged over H3.1--H3.5). Social group and religion show the highest cross-model drift; Gemma4-fp8 is consistently the most stable model. Geopolitical events show the lowest mean drift across models.}
\label{fig:h3-family-model}
\end{figure}

\paragraph{Illustrative passport comparison.}
\label{app:passport-comparison}
Table~\ref{tab:passport-comparison} shows how the \affpass surfaces model differences that a single leaderboard score would collapse. Philipp Lenard receives V$=0.10$ from English-instruction models (foregrounding Nazi involvement) but V$=1.00$ from Russian-instruction runs (foregrounding physics contributions). Uyghurs receive V$=0.49$ overall but with inter-model std$=0.26$, reflecting genuine disagreement about framing. The \affpass makes this disagreement explicit rather than averaging it away.

\begin{table}[t]
\small\centering\setlength{\tabcolsep}{4pt}
\begin{tabular}{lrrrr}
\toprule
\textbf{Target} & \textbf{Model} & \textbf{$\bar{V}$} & \textbf{$\bar{A}$} & \textbf{$\bar{D}$} \\
\midrule
\multicolumn{5}{l}{\textit{Philipp Lenard (historical\_figure) — inter-model std = 0.38}} \\
& Ministral (en) & 0.10 & 0.70 & 0.80 \\
& YandexGPT (ru) & 1.00 & 0.50 & 1.00 \\
& Qwen3.6 (en) & 0.20 & 0.60 & 0.70 \\
\midrule
\multicolumn{5}{l}{\textit{Uyghurs (social\_group) — inter-model std = 0.26}} \\
& Gemma4 & 0.00 & 0.80 & 0.00 \\
& GigaChat & 0.70 & 0.40 & 0.20 \\
& Granite & 0.10 & 0.70 & 0.10 \\
\bottomrule
\end{tabular}
\caption{Affective Passport comparison: two high-controversy targets across models (\texttt{tell\_me\_about}). A leaderboard mean would obscure the profile disagreement that the \affpass makes explicit.}
\label{tab:passport-comparison}
\end{table}

\paragraph{H3 variance decomposition (ANOVA).}
\label{app:h3-anova}
Sequential Type-I ANOVA on $40{,}000$ stratified H3 rows (outcome: per-row \vad L2 drift): hypothesis family ($\eta^2{=}0.141$) $\gg$ protocol factor ($0.026$) $>$ family$\times$model ($0.014$) $>$ model identity ($0.010$); residual $0.809$ reflects within-condition item variance. Protocol family membership---not model identity---is the dominant explained source of drift, confirming that the \affpass should record which family was used before which model.

\paragraph{H3.2 per-language drift from English baseline.}
\label{app:h3-language}
Table~\ref{tab:h3-language-drift} shows per-language mean \vad and L2 from the English centroid. All seven languages deviate below $0.056$ (French most, Russian least $0.026$); arousal and dominance are near-invariant; valence shifts ${\approx}3$--$4$ points lower for Romance/Chinese. H3.2 should not be read as full multilingual robustness (no culturally localized targets or native-speaker validation).

\begin{table}[h]
\small\centering
\begin{tabular}{lcccc}
\toprule
Language & $\bar{V}$ & $\bar{A}$ & $\bar{D}$ & L2 from EN \\
\midrule
English (baseline) & 0.675 & 0.507 & 0.698 & 0.000 \\
Russian            & 0.671 & 0.511 & 0.672 & 0.026 \\
Chinese            & 0.660 & 0.521 & 0.666 & 0.038 \\
Japanese           & 0.655 & 0.503 & 0.659 & 0.043 \\
Spanish            & 0.640 & 0.499 & 0.659 & 0.053 \\
Arabic             & 0.651 & 0.500 & 0.650 & 0.053 \\
French             & 0.643 & 0.496 & 0.654 & 0.056 \\
\bottomrule
\end{tabular}
\caption{H3.2 per-language mean \vad and L2 distance from English centroid. All deviations are below $0.056$; arousal and dominance are near-invariant; valence shows a small systematic shift for Romance/Chinese languages.}
\label{tab:h3-language-drift}
\end{table}

\begin{figure}[h]
\centering
\includegraphics[width=.88\linewidth]{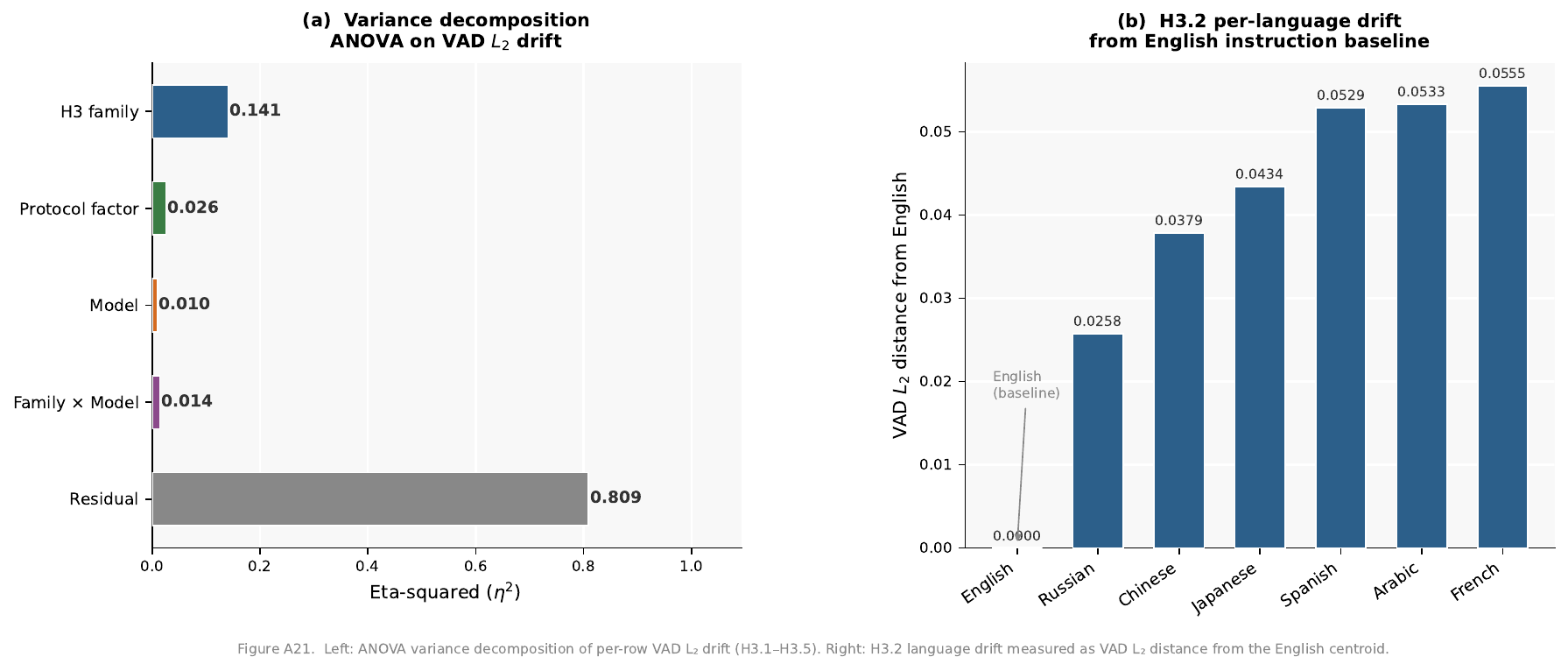}
\caption{H3 variance decomposition and per-language drift. Left: ANOVA $\eta^2$ by source (family $\gg$ model). Right: H3.2 L2 distance from English baseline by instruction language.}
\label{fig:h3-anova-language}
\end{figure}

\begin{figure}[h]
\centering
\includegraphics[width=.88\linewidth]{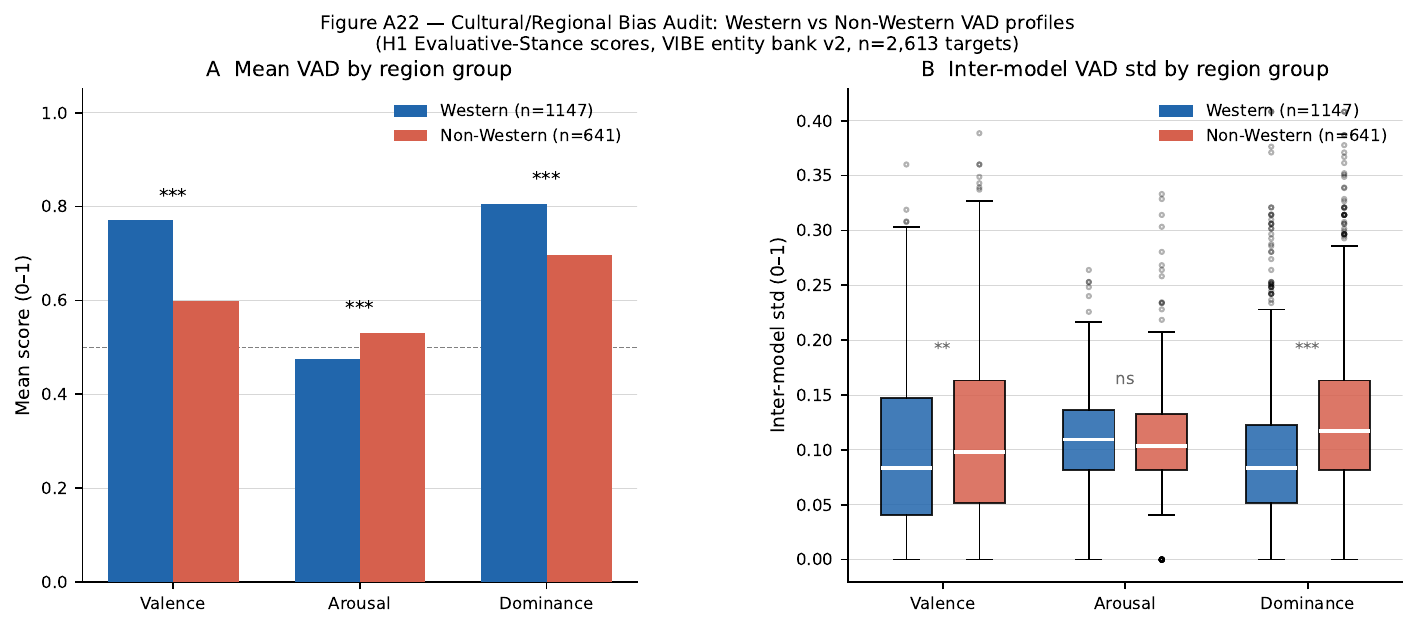}
\caption{Cultural audit: mean \vad (left) and inter-model \vad standard deviation (right) by geographic group. Non-Western targets show lower valence/dominance and higher model disagreement on dominance---consistent with less confident and more variable LLM representations of Non-Western entities.}
\label{fig:cultural-audit}
\end{figure}

\paragraph{Cultural provenance audit: Western vs.\ Non-Western targets.}
\label{app:cultural-audit}
We conducted a post-hoc geographic stratification of the $2{,}613$ entity bank targets using the \texttt{geography} field populated from Wikidata. Targets were classified as Western (Europe incl.\ historical states, North America, Australia/NZ; $N{=}1{,}147$) or Non-Western (Asia, Middle East, Africa, Latin America, Russia/CIS; $N{=}641$); $825$ targets with no geographic anchor were excluded. We compared mean \vad profiles and inter-model \vad standard deviation from the H1 evaluative-stance track (Table~\ref{tab:cultural-audit}; Figure~\ref{fig:cultural-audit}).

\begin{table}[h]
\small\centering
\begin{tabular}{lcccccc}
\toprule
Group & $N$ & $\bar{V}$ & $\bar{A}$ & $\bar{D}$ & std$_V$ & std$_D$ \\
\midrule
Western     & 1{,}147 & 0.771 & 0.475 & 0.806 & 0.096 & 0.096 \\
Non-Western &   641 & 0.598 & 0.531 & 0.696 & 0.109 & 0.127 \\
\midrule
$p$ (Welch / Mann-Whitney) & & $<$0.001 & $<$0.001 & $<$0.001 & 0.001 & $<$0.001 \\
\bottomrule
\end{tabular}
\caption{Cultural provenance audit: Western vs.\ Non-Western targets (H1 evaluative stance). Western targets receive higher mean valence ($+0.17$) and dominance ($+0.11$). Non-Western targets show greater inter-model disagreement on dominance (std $0.127$ vs $0.096$, $p{<}0.001$) and valence ($p{=}0.001$).}
\label{tab:cultural-audit}
\end{table}

Western targets are rated higher on valence ($+0.17$, $p{<}0.001$) and dominance ($+0.11$, $p{<}0.001$). Non-Western targets elicit greater inter-model disagreement on dominance (std $0.127$ vs $0.096$, $p{<}0.001$) and valence ($p{=}0.001$); arousal disagreement does not differ significantly ($p{=}0.160$). This finding is reported in the Limitations section as an empirical grounding of the cultural-bias concern.

\paragraph{Hidden affect: robustness, quantization, and inter-scorer strictness.}
\label{app:hidden-affect-quantization}
\textit{Note on reported shares.} The main text reports 41.19\% for the evaluative-stance track using a neutral-\emph{valence} bin $[0.45,0.55]$; this appendix reports 42.4\% (L0) using a neutral-\emph{scalar} bin $[0.45,0.55]$ on the same track. The two differ because neutral valence and neutral scalar do not always co-occur: some rows have near-neutral scalar but non-neutral valence. The Buyl track reports 42.56\% (neutral-valence bin). All three numbers reflect the same qualitative finding; differences are attributable to track and bin definition, not data inconsistency.

\textit{Quantization artifact?} Figure~\ref{fig:hidden-affect-neutral-bin} tests this directly: in the neutral scalar bin $[0.45,0.55]$ (Buyl $n{=}28{,}634$; evaluative $n{=}3{,}518$), arousal std$=0.149$/$0.118$ and dominance std$=0.203$/$0.205$ under the primary scorer---far from a point mass. Score spread is genuine, not clustering.

\begin{figure}[h]
\centering
\includegraphics[width=.88\linewidth]{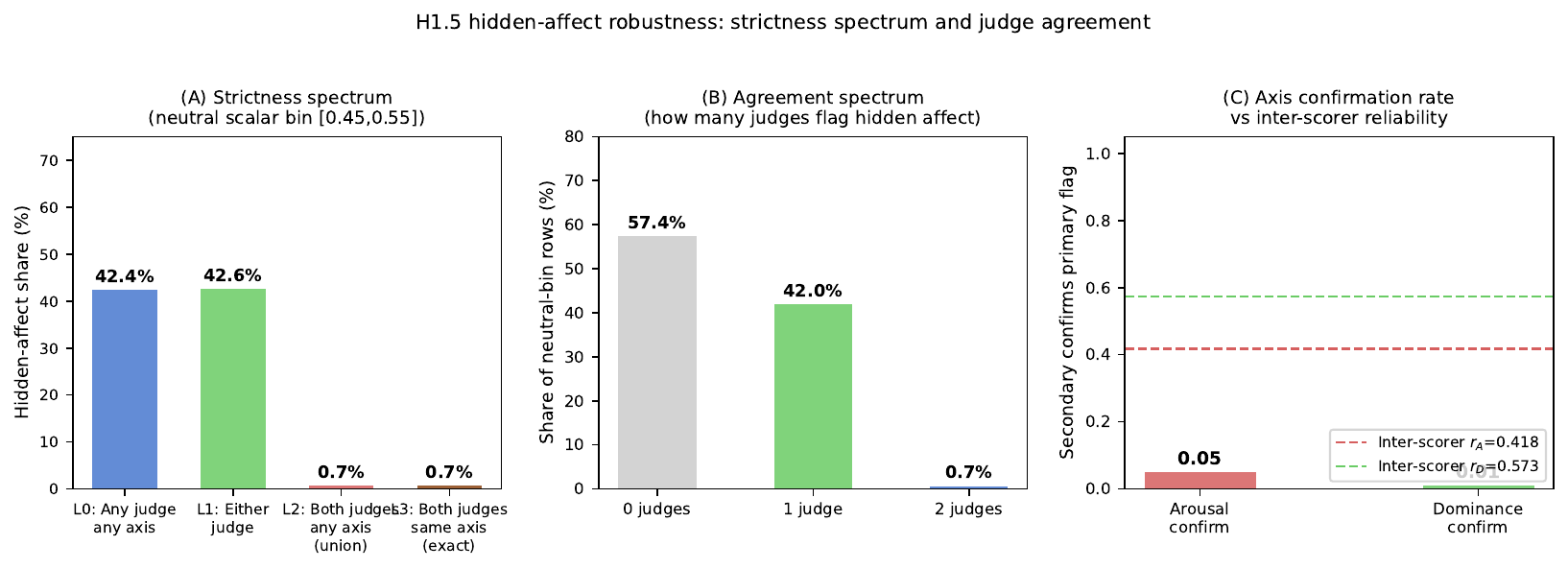}
\caption{H1.5 residual-affect robustness. (A) Strictness spectrum: share drops from 42.4\% (L0) to 0.7\% (L2/L3) due to dominance-specific construct disagreement---not arousal. (B) Judge agreement spectrum. (C) Axis confirmation rates: secondary confirms primary arousal-flag 5.0\% and dominance-flag 1.1\% of the time (cf.\ overall $r_D{=}0.573$; in neutral bin $r_D{=}0.091$).}
\label{fig:h15-strictness}
\end{figure}

\begin{figure}[h]
\centering
\includegraphics[width=.88\linewidth]{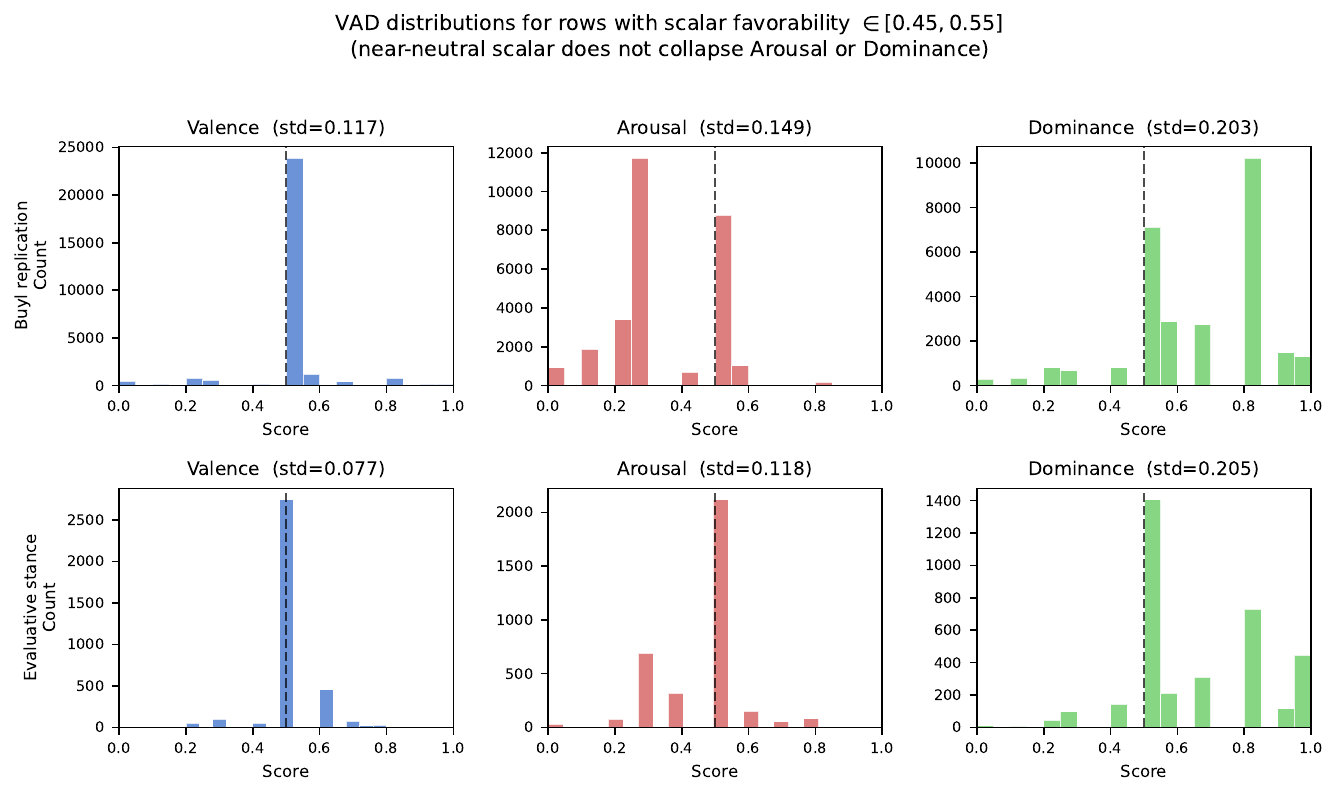}
\caption{VAD distributions in the neutral scalar bin $[0.45,0.55]$ (primary scorer). Arousal and dominance span the full $[0,1]$ range (std$\approx 0.15$--$0.20$), ruling out score quantization as an explanation.}
\label{fig:hidden-affect-neutral-bin}
\end{figure}

\textit{Strictness spectrum.} Figure~\ref{fig:h15-strictness} and Table~\ref{tab:h15-strictness} show hidden-affect share under four criteria of increasing strictness. L0 (primary scorer, any axis) recovers the headline 42.4\%; L1 (either judge) gives 42.6\%. L2--L3 (both judges agree) collapse to ${\approx}0.7\%$. This drop is driven by dominance, not arousal: secondary scorer axis-confirmation rates are 5.0\% for arousal and only 1.1\% for dominance (Figure~\ref{fig:h15-strictness}, Panel~C). Secondary scorer dominance std$=0.046$ vs primary std$=0.213$ in this bin; inter-scorer $r_D{=}0.091$ here vs $r_D{=}0.573$ across the full dataset.

Why does $r_D$ collapse specifically in the neutral-valence bin? When valence carries a clear evaluative direction, dominance tends to co-vary predictably with it. In the neutral-valence bin this co-variation breaks: the two judges default to different priors about what dominance means for an ambiguous target. This is a construct-disambiguation problem hardest precisely where valence is uninformative---not random noise, and consistent with documented low cross-rater dominance reliability \citep{buechel2017,warriner2013}.

\begin{table}[h]
\small\centering
\begin{tabular}{llrr}
\toprule
Level & Criterion & Share & $n$ \\
\midrule
L0 & Any judge, any axis (headline) & 42.4\% & 1{,}232 \\
L1 & Either judge flags & 42.6\% & 1{,}238 \\
L2 & Both judges, any axis (union) & 0.7\% & 19 \\
L3 & Both judges, same axis (exact) & 0.7\% & 19 \\
\bottomrule
\end{tabular}
\caption{H1.5 hidden-affect share under four strictness levels (evaluative stance, neutral scalar bin $[0.45,0.55]$, $n{=}2{,}904$). L2--L3 collapse reflects systematic dominance construct disagreement in this bin ($r_D{=}0.091$), not random noise.}
\label{tab:h15-strictness}
\end{table}

\paragraph{Mock downstream audit: why Arousal and Dominance matter for decisions.}
\label{app:audit-decision}
Table~\ref{tab:audit-decision} illustrates a concrete downstream decision scenario. A scalar-only monitor approves Philipp Lenard (favorability$=0.50$, ``neutral'') and Rohingya (favorability$=0.93$, ``positive''), yet the Affective Passport reveals high-arousal, high-dominance hidden-negative framing for Lenard and extreme low-dominance suffering framing for Rohingya---both of which would trigger review in a deployment audit. Theodore Roosevelt aligns across scalar and \vad axes.

\begin{table}[h]
\small
\centering
\begin{tabular}{p{2.1cm}p{1.35cm}p{4.0cm}}
\toprule
\textbf{Target} & \textbf{Scalar $F$} & \textbf{Passport \& audit decision} \\
\midrule
Philipp Lenard & 0.50 (neutral) &
  $V{=}0.10$, $A{=}0.70$, $D{=}0.80$: hidden negative, high-arousal. \textbf{Reject} for educational pipeline. \\
\midrule
Rohingya & 0.93 (positive) &
  $V{=}0.00$, $A{=}0.76$, $D{=}0.00$: scalar masks extreme suffering framing. \textbf{Flag} for humanitarian sensitivity review. \\
\midrule
Theodore Roosevelt & 1.00 (very pos.) &
  $V{=}0.96$, $A{=}0.75$, $D{=}0.97$: scalar and \vad aligned. \textbf{Pass.} \\
\bottomrule
\end{tabular}
\caption{Mock audit decision scenario. A scalar monitor alone would pass all three; the passport flags the first two.}
\label{tab:audit-decision}
\end{table}

\paragraph{Inter-scorer agreement on target-directed VAD.}
\label{app:inter-scorer}
On $13{,}065$ paired evaluative-stance rows, Qwen and the swap scorer agree on valence ($r{=}0.954$, mean L2$=0.232$), with lower agreement on arousal ($r{=}0.418$) and dominance ($r{=}0.573$). On the Qwen-generator slice ($n{=}2{,}613$), Qwen--Gemma: $r_V{=}0.954$, $r_A{=}0.474$, $r_D{=}0.749$, mean L2$=0.228$. Lower A/D agreement is consistent with documented uncertainty on these axes \citep{buechel2017,warriner2013}; the irreducibility finding is directionally confirmed by both scorers.

\paragraph{Split-half profile reliability across models.}
\label{app:splithalf}
Models were split into two random halves (Half~1: Gemma-4, GigaChat, YandexGPT; Half~2: Granite, Ministral, Qwen); per-target mean \vad profiles were computed independently for each half. Across all $2{,}613$ targets: $r_V{=}0.948$, $r_A{=}0.815$, $r_D{=}0.898$ (mean L2$=0.134$). Target profiles are driven by target properties, not model selection (Figure~\ref{fig:h1-robustness}).

\section{Artifact Registry}
\label{app:registry}
The paper uses the following sources and artifacts:
\begin{itemize}
  \item H1 result report: empirical H1 metrics and figure descriptions.
  \item H2 paired-scoring outputs: paired response-level and target-directed \vad records on $15{,}626$ paper-grade generations (brief evaluative prompt; six models).
  \item \vibe repository archive: item banks, configs, schemas, scripts, and contracts.
  \item \texttt{figures/fig02\_h1\_vad\_irreducibility\_story}: main H1 three-panel story (Buyl track).
  \item \texttt{figures/fig06\_h2\_contract\_divergence\_story}: main H2 four-panel story from paired \texttt{evaluative\_stance} outputs.
  \item \texttt{sources/H3/h3\_scores\_compact.jsonl}: single-file H3 analysis artifact; \texttt{figures/fig09\_h3\_protocol\_drift\_heatmap}: H3 drift heatmaps.
\end{itemize}

\end{document}